\documentclass[10pt,journal,compsoc]{IEEEtran}
\usepackage{amsmath,amsfonts}
\usepackage{algorithmic}
\usepackage{algorithm}
\usepackage{array}
\usepackage{url}
\usepackage{graphicx}
\usepackage{booktabs}
\usepackage{multirow}
\usepackage{hhline}
\usepackage{bm}
\usepackage{adjustbox}
\ifCLASSOPTIONcompsoc
\usepackage[nocompress]{cite}
\else
\usepackage{cite}
\fi
\ifCLASSINFOpdf
\else
\fi
	\usepackage{times}
	\usepackage{epsfig}
	\usepackage{amssymb}
	\usepackage{bm}
	\usepackage{bbding}
	\usepackage{multirow}
	\usepackage{booktabs,graphicx}
	\usepackage{animate}
	\usepackage{tabu}
	\usepackage{xspace}
	\usepackage{pifont}
	\usepackage[pagebackref=true,breaklinks=true,letterpaper=true,colorlinks,bookmarks=false]{hyperref}
	
	\makeatletter
	\DeclareRobustCommand\onedot{\futurelet\@let@token\@onedot}
	\def\@onedot{\ifx\@let@token.\else.\null\fi\xspace}
	
	\def\eg{\emph{e.g}\onedot} 
	\def\ie{\emph{i.e}\onedot}

	\def\etal{\emph{et al}\onedot}
	\makeatother
	
\begin{document}
		\title{BasicAVSR: Arbitrary-Scale Video Super-Resolution via Image Priors and Enhanced Motion Compensation}
		
		\author{Wei Shang\thanks{Wei Shang is with the Faculty of Computing, Harbin Institute of Technology, Harbin, China. (E-mail: csweishang@gmail.com)}, 
			Wanying Zhang\thanks{Wanying Zhang is with the Faculty of Computing, Harbin Institute of Technology, Harbin, China. (E-mail: swzwanying@gmail.com)}, 
			Shuhang Gu\thanks{Shuhang Gu u is with the School of Computer Science and Engineering, University of Electronic Science and Technology of China, Chengdu, China. (E-mail: shuhanggu@gmail.com)}, 
			Pengfei Zhu\thanks{Pengfei Zhu is with the Tianjin Key Laboratory of Machine Learning, College of Intelligence and Computing, Tianjin University, Tianjin, China. (E-mail: zhupengfei@tju.edu.cn)}, 
			Qinghua Hu\thanks{Qinghua Hu is with the Tianjin Key Laboratory of Machine Learning, College of Intelligence and Computing, Tianjin University, Tianjin, China. (E-mail: huqinghua@tju.edu.cn)},
			and Dongwei Ren\thanks{Dongwei Ren is with the Tianjin Key Laboratory of Machine Learning, College of Intelligence and Computing, Tianjin University, Tianjin, China. (Corresponding Author. E-mail: rendw@tju.edu.cn)}}
		

	\markboth{Journal of \LaTeX\ Class Files,~Vol.~14, No.~8, August~2024}%
	{Shang \MakeLowercase{\textit{et al.}}: BasicAVSR: Arbitrary-Scale Video Super-Resolution via Image Priors and Enhanced Motion Compensation}



	
	\IEEEtitleabstractindextext{%
		%\begin{abstract}
		%The abstract goes here.
		%\end{abstract}
		%
		%% Note that keywords are not normally used for peerreview papers.
		%\begin{IEEEkeywords}
		%Computer Society, IEEE, IEEEtran, journal, \LaTeX, paper, template.
		%\end{IEEEkeywords}
		\begin{abstract}
			Arbitrary-scale video super-resolution (AVSR) aims to enhance the resolution of video frames, potentially at various scaling factors, which presents several challenges regarding spatial detail reproduction, temporal consistency, and computational complexity. In this paper, we propose a strong baseline BasicAVSR for AVSR by integrating four key components: 1) adaptive multi-scale frequency priors generated from image Laplacian pyramids, 2) a flow-guided propagation unit to aggregate spatiotemporal information from adjacent frames, 3) a second-order motion compensation unit for more accurate spatial alignment of adjacent frames, and 4) a hyper-upsampling unit to generate scale-aware and content-independent upsampling kernels.
			To meet diverse application demands, we instantiate three propagation variants: (i) a unidirectional RNN unit for strictly online inference, (ii) a unidirectional RNN unit empowered with a limited lookahead that tolerates a small output delay, and (iii) a bidirectional RNN unit designed for offline tasks where computational resources are less constrained. 
			Experimental results demonstrate the effectiveness and adaptability of our model across these different scenarios.
			Through extensive experiments, we show that BasicAVSR significantly outperforms existing methods in terms of super-resolution quality, generalization ability, and inference speed. Our work not only advances the state-of-the-art in AVSR but also extends its core components to multiple frameworks for diverse scenarios. The code is available at {\url{https://github.com/shangwei5/BasicAVSR}}.
		\end{abstract}
		
		\begin{IEEEkeywords}
			Arbitrary-scale video super-resolution, frequency priors, motion compensation.
		\end{IEEEkeywords}
	}

	\maketitle

	\IEEEdisplaynontitleabstractindextext

	% \fi
	\IEEEpeerreviewmaketitle

	\IEEEraisesectionheading{\section{Introduction}}
	\label{sec:intro}

	\IEEEPARstart{T}{he} evolutionary and developmental processes of our visual systems have presumably been shaped by continuous visual data~\cite{wandell1995foundations}. Yet, how to acquire and represent a natural scene as a continuous signal remains wide open. This difficulty stems from two main factors. The first is the physical limitations of digital imaging devices, including sensor size and density, optical diffraction, lens quality, electrical noise, and processing power. The second is the inherent complexities of natural scenes, characterized by their wide and deep frequencies, which pose significant challenges for applying the Nyquist–Shannon sampling~\cite{oppenheim1997signals} and compressed sensing~\cite{donoho2006compressed} theories to accurately reconstruct continuous signals from discrete samples. Consequently, natural scenes are predominantly represented as discrete pixel arrays, often with limited resolution.
	
	Super-resolution (SR) provides an effective means of enhancing the resolution of low-resolution (LR) images and videos~\cite{irani1991improving,shechtman2005space}. Early deep learning-based SR methods~\cite{dong2014learning,shi2016real,lim2017enhanced,zhang2018residual} focus on fixed integer scaling factors (\eg, $\times 4$ and $\times 8$), each corresponding to an independent convolutional neural network (CNN). This limits their applicability in real-world scenarios, where varying scaling requirements are common. From the human vision perspective, users may want to continuously zoom in on images and videos to arbitrary scales using the two-finger pinch-zoom feature on mobile devices as a natural form of human-computer interaction.  From the machine vision perspective, different applications (such as computer-aided diagnosis, remote sensing, and video surveillance) may require different scaling factors to zoom in on different levels of detail for optimal analysis and decision-making.
	
	Recently, arbitrary-scale image SR (AISR) has gained significant attention due to its capability of upsampling LR images to arbitrary high-resolution (HR) using a single model. 
	Contemporary AISR methods can be categorized into four classes based on how arbitrary-scale upsampling is performed: interpolation-based methods~\cite{kim2016accurate,behjati2021overnet}, learnable adaptive filter-based methods~\cite{hu2019meta,wang2021learning,wang2023deep}, implicit neural representation-based methods~\cite{chen2021learning,lee2022local, chen2023cascaded}, and Gaussian splatting-based methods~\cite{hu2025gaussiansr,peng2025pixel}. These algorithms face several limitations, including quality degradation at high (and possibly integer) scales~\cite{hu2019meta,chen2021learning,wang2021learning}, high computational complexity~\cite{lee2022local, chen2023cascaded}, 
	and difficulty in generalizing across unseen scales and degradation models~\cite{hu2019meta,chen2021learning,lee2022local, chen2023cascaded}, as well as temporal inconsistency in video SR.

	Compared to AISR,  arbitrary-scale video SR (AVSR) is significantly more challenging due to the added time dimension.
	Existing AVSR methods~\cite{chen2022videoinr,chen2023motif,kim2025bf} rely primarily on conditional neural radiance fields~\cite{mildenhall2020nerf} as continuous signal representations. Due to the high computational demands during training and inference, only two adjacent frames are used for spatiotemporal modeling, which is bound to be suboptimal. Another method~\cite{li2024savsr} adopts an upsampling kernel, thereby avoiding the drawbacks of conditional neural radiance fields. However, treating scale factors as priors can not provide guidance related to image content. The sliding-window-based bidirectional RNN utilized in this method has limitations in modeling long-sequence motions and is computationally inefficient. Moreover, current AVSR approaches rely on optical flows for inter-frame alignment; yet, imprecise optical flow estimation degrades model performance.
	
\begin{figure*}[!t]\footnotesize
		\centering
		\setlength{\abovecaptionskip}{3pt} 
		\setlength{\belowcaptionskip}{0pt}
		\begin{tabular}{cccccc}
			\includegraphics[width=\linewidth]{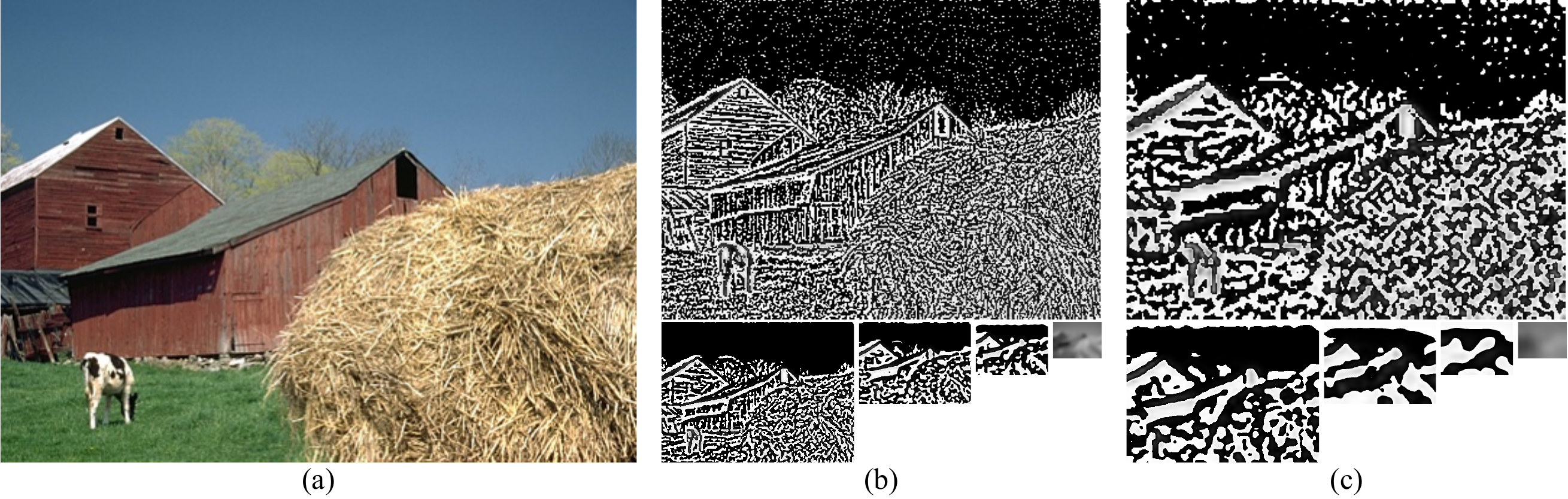}\\
		\end{tabular}
		\caption{
			Visualization of Laplacian pyramid decomposition under different input resolutions. (a) Original image. (b) Laplacian pyramid visualization (high-resolution input). (c) Laplacian pyramid visualization (low-resolution input). 
		}
		\label{fig:texture}
	\end{figure*}

	In this work, we aim for AVSR with the goal of reproducing faithful spatial detail and maintaining coherent temporal consistency at low computational complexity. 
	We describe a strong baseline, which we name BasicAVSR, by identifying and combining four variants of elementary building blocks~\cite{chan2022basicvsr++,burt1983laplacian}: 1) multi-scale frequency priors, 2) a flow-guided propagation unit, 3) a second-order motion compensation unit, and 4) a hyper-upsampling unit. 
	BasicAVSR is grounded in scale-space theory~\cite{lindeberg2013scale} in computer vision and image processing, which suggests that human perception and interpretation of real-world structures and textures are scale-dependent. 
	As shown in Fig.~\ref{fig:texture}, multi-scale frequency priors offer rich frequency-domain information, which is beneficial for restoring detailed textures and structures in VSR. The same image reveals distinct frequency-band spatial differences across different input resolutions. Accurately characterizing these multi-scale image frequency bands is highly valuable for AVSR. 
	We extract frequency-domain priors adaptively from the Laplacian pyramid decomposition of each frame. These priors effectively distinguish structures and textures across scales and capture mid-level visual concepts tied to image layout~\cite{fu2023dreamsim,zhou2023dichotomous}.
	The flow-guided propagation unit captures long-term spatiotemporal dependencies from adjacent frames, thereby improving the temporal consistency of the video sequence. The second-order motion compensation unit refines the spatial alignment of adjacent frames and searches for regions with similar image content near the initially estimated motion offset, thereby achieving more accurate motion estimation. The hyper-upsampling unit trains a hyper-network~\cite{ha2017hypernetworks} that takes scale-relevant parameters as input to generate content-independent upsampling kernels, enabling pre-computation to accelerate inference speed. 
	
	%Overall, these components work synergistically to achieve high-quality VSR with improved efficiency and effectiveness.

	%

	This work is primarily presented as a conference paper~\cite{shang2024arbitrary}, upon which this manuscript has made two major improvements, i.e., a more direct way of generating multi-scale priors is introduced to eliminate the need for extra pre-trained networks, and a motion compensation mechanism is introduced to enhance the accuracy of video frames alignment. In addition, the method is extended to multiple application scenarios, including online, offline, and quasi-online settings. To sum up, the main contributions of this work include:
	
	\begin{itemize}
		\item 
		A strong baseline, BasicAVSR, that is a nontrivial combination of four variants of elementary building blocks in literature~\cite{chan2022basicvsr++,burt1983laplacian},
		\item A method for obtaining image-content-related priors based on the Laplacian pyramid and a motion compensation strategy that searches for similar content near the initial motion estimation, 
		\item An extension for adapting core modules and strategies to online, offline, and quasi-online scenarios, giving rise to three variants: unidirectional RNN, bidirectional RNN, and unidirectional RNN with lookahead, and 
		\item  A comprehensive experiment that shows BasicAVSR significantly surpasses competing methods in terms of SR quality on different test sets, generalization ability to unseen scales and degradation models, as well as inference speed.
	\end{itemize}

	%-------------------------------------------------------------------------
	\section{Related Work}
	
	In this section, we review key components of VSR, upsampling modules for AISR and AVSR, and natural scene priors employed in SR.
	\subsection{Key Components of VSR}
	Kappeler \etal~\cite{kappeler2016video} pioneered CNN-based approaches for VSR, emphasizing two key components: feature alignment and aggregation. Subsequent studies have focused on enhancing these components. EDVR~\cite{wang2019edvr} introduced pyramid deformable alignment and spatiotemporal attention for feature alignment and aggregation.
	BasicVSR~\cite{chan2021basicvsr} and BasicVSR$++$~\cite{chan2022basicvsr++}
	employ an optical flow-based module to estimate motion correspondence between neighboring frames for feature alignment and a bidirectional propagation module to aggregate spatiotemporal information from previous and future frames, which set the VSR performance record at that time.
	RVRT~\cite{liang2022recurrent} enhanced VSR performance by utilizing a recurrent video restoration Transformer with guided deformable attention albeit at the expense of substantially increased computational complexity.
	Additionally, VideoINR~\cite{chen2022videoinr}, MoTIF~\cite{chen2023motif}, and BF-STVSR~\cite{kim2025bf}  integrated VSR with video frame interpolation, which achieved limited success due to the ill-posedness of the task.
	In our work, we combine a flow-guided propagation unit and a second-order motion compensation unit to extract, align, and aggregate spatiotemporal features from adjacent frames, while keeping computational complexity manageable.
	
	\subsection{Upsampling Modules for AISR and AVSR}
	
	Compared to fixed-scale SR methods ~\cite{dong2014learning,shi2016real,lim2017enhanced,zhang2018residual,liang2021swinir}, upsampling plays a more crucial role in AISR and AVSR.
	Besides direct interpolation-based upsampling~\cite{kim2016accurate,behjati2021overnet}, learnable adaptive filter-based upsampling, implicit neural representation-based upsampling, and Gaussian splatting-based upsampling are commonly used. Meta-SR~\cite{hu2019meta} was the pioneer in AISR, dynamically predicting the upsampling kernels using a single model.
	ArbSR~\cite{wang2021learning} introduced a scale-aware upsampling layer compatible with fixed-scale SR methods.
	EQSR~\cite{wang2023deep} proposed a bilateral encoding of both scale-aware and content-dependent features during upsampling.
	Inspired by the success of implicit neural representations in computer graphics~\cite{michalkiewicz2019implicit,peng2020convolutional}, this approach has also been applied to AISR and AVSR.
	For instance, LIIF~\cite{chen2021learning} predicts the RGB values of HR pixels using the coordinates of LR pixels along with their neighboring features as inputs.
	LTE~\cite{lee2022local} captures more fine detail with a local texture estimator, and
	CLIT~\cite{chen2023cascaded} enhances representation expressiveness 
	with cross-scale attention and multi-scale reconstruction.
	OPE~\cite{song2023ope} introduced orthogonal position encoding for efficient upsampling. 
	CiaoSR~\cite{cao2023ciaosr} proposed an attention-based weight ensemble algorithm for feature aggregation in a large receptive field.
	Recently, 2D Gaussian Splatting has shown great potential in image processing~\cite{hu2025gaussiansr,peng2025pixel}. Unlike traditional methods that treat pixels as discrete points, Gaussian splatting-based upsampling represents each pixel as a continuous Gaussian field. By rendering mutually stacked Gaussian fields, the encoded features are simultaneously refined and upsampled, which establishes long-range dependencies and enhances representation ability.
	
	Existing AVSR methods~\cite{chen2022videoinr,chen2023motif, kim2025bf} also use implicit neural representations but are constrained to modeling spatiotemporal relationships between only two adjacent frames due to the high computational costs involved. 
	In contrast to these approaches, SAVSR~\cite{li2024savsr} introduces a dual-branch upsampling architecture. It adaptively adjusts network weights based on spatiotemporal features and upsampling scales. However, the computational process of both branches relies on input features, which means pre-calculation is not feasible in this architecture.
	The proposed BasicAVSR addresses this limitation by employing a lightweight hyper-upsampling unit to predict scale-aware and content-independent upsampling kernels, allowing for pre-computation to speed up inference.
	
	\subsection{Natural Scene Priors for SR}
	The history of SR, or more general low-level vision, is closely tied to the development of natural scene priors. Commonly used priors in SR include the smoothness prior~\cite{chambolle2004algorithm}, sparsity prior~\cite{mairal2014sparse}, self-similarity prior~\cite{glasner2009super}, edge/gradient prior~\cite{he2011single}, deep architectural prior~\cite{ulyanov2018deep}, temporal consistency prior~\cite{caballero2017real}, motion prior~\cite{tao2017detail,shang2023joint}, and perceptual prior~\cite{wang2018recovering}. 
	In the subfield of AISR and AVSR, the scaling factor-based priors have exclusively been leveraged as adaptive convolution conditions~\cite{wang2021learning,fu2021residual,wang2023deep,li2024savsr}. However, single scaling factors are limited in providing rich texture priors that vary with the video content.
	In this paper, we introduce multi-scale frequency priors. Using Laplacian pyramid decomposition, we effectively distinguish between high- and low-frequency information of images at varying locations and scales. We demonstrate its effectiveness in enhancing AVSR.
	
	\begin{figure*}[!t]\footnotesize
		\centering
		\setlength{\abovecaptionskip}{3pt} 
		\setlength{\belowcaptionskip}{0pt}
		\begin{tabular}{cccccc}
			\includegraphics[width=\linewidth]{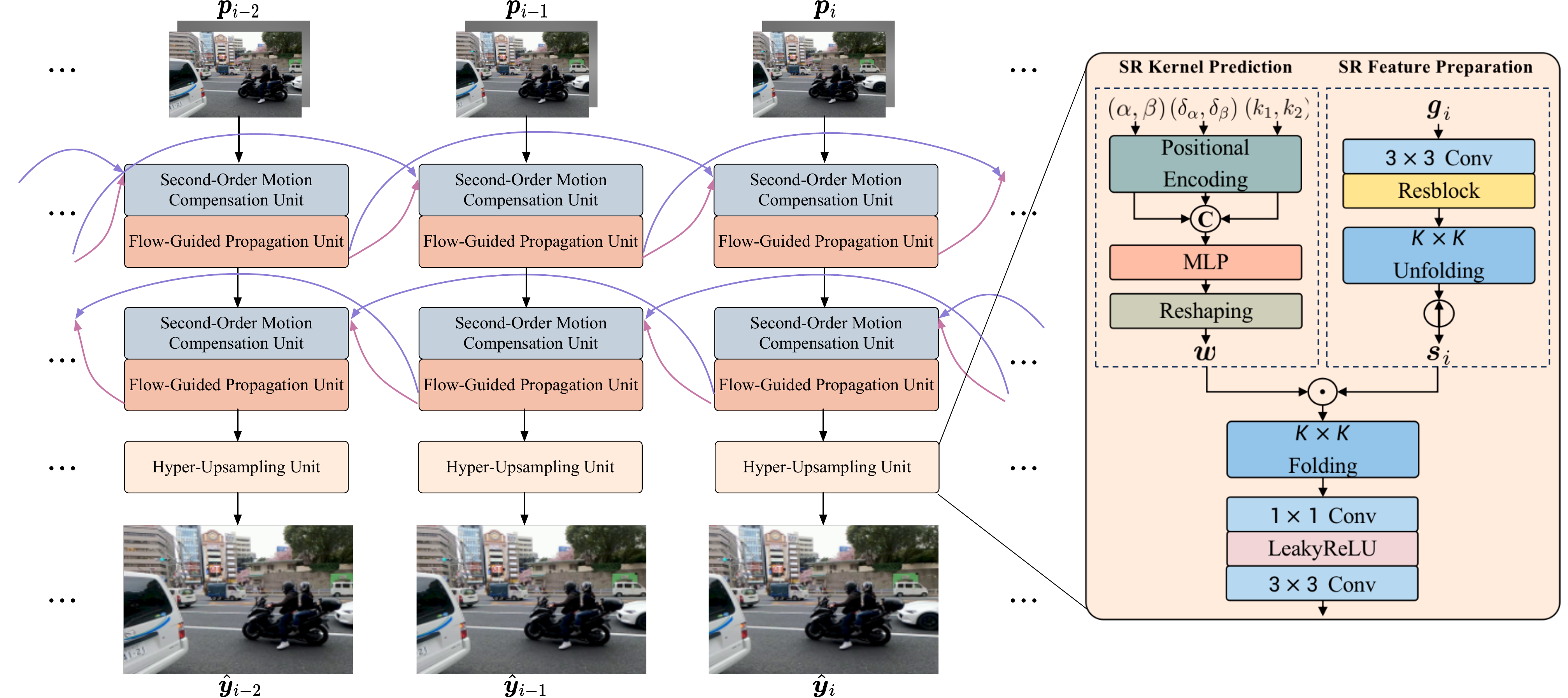}\\
		\end{tabular}
		\caption{System diagram of BasicAVSR, which reconstructs an arbitrary-scale HR video $\hat{\bm y}$ from an LR video input $\bm x$. BasicAVSR is composed of four variants of elementary building blocks: 1) multi-scale frequency priors to provide scale-specific pixel-level priors for AVSR by replacing all instances of $\bm x$ with the multi-scale frequency prior $\bm p$ (see the detailed text description in Sec.~\ref{subsec:st}), 2) a flow-guided propagation unit to aggregate features from adjacent frames, 3) a second-order motion compensation unit to mitigate misalignment in backward warping (see also Fig.~\ref{fig:local}), and 4) a hyper-upsampling unit to prepare SR features and predict SR kernels for HR frame reconstruction.
		}
		\label{fig:framework}
	\end{figure*}
	
	\section{Proposed Method: BasicAVSR}
	%\subsection{Overall Framework of BasicAVSR}
	Given an LR video sequence $\bm{x} = \{\bm{x}_{i}\}_{i=0}^T$, where $\bm x_{i}\in\mathbb{R}^{H\times W}$ is the $i$-th frame, and $H$ and $W$ are the frame height and width, respectively, the goal of the proposed BasicAVSR is to reconstruct an HR video sequence $\hat{\bm{y}} = \{\hat{\bm{y}}_{i}\}_{i=0}^T$ with $\hat{\bm y}_{i} \in \mathbb{R}^{(\alpha H)\times (\beta W)}$, where $\alpha,\beta\ge 1$ are two user-specified scaling factors. 
	Our baseline BasicAVSR consists of four variants of basic building blocks: 1) multi-scale frequency priors to inject content-dependent pixel-level cues to guide restoration, 2) a flow-guided propagation unit to aggregate spatiotemporal information from adjacent frames, 3) a second-order motion compensation unit to perform accurate sub-pixel alignment through a coarse-to-fine refinement strategy, and 4) a hyper-upsampling unit to generate scale-specific kernels that can be pre-computed to enable AVSR. 
	We next detail each component using bidirectional RNN as an example (system diagram in Fig.~\ref{fig:framework}). Finally, the two alternative propagation variants are presented in Sec.~\ref{sec:arch_exten}.

	\subsection{Multi-Scale Frequency Priors for AVSR}\label{subsec:st}
	Accurately characterizing image structure and texture at multiple scales is crucial for the task of AVSR. Fortunately, the scale-space theory in computer vision and image processing~\cite{koenderink1984structure,lindeberg2013scale} provides an elegant theoretical framework for this purpose. 
	Since different frequency bands of an image can present the structure and texture of the image, in this work, we adopt multi-scale frequency priors derived from Laplacian pyramid decomposition as an alternative to the VGG-based features used in our previous work~\cite{shang2024arbitrary}. 
	This approach breaks down an image into multiple layers of different frequency bands. The resulting pyramid consists of several levels, each representing the image at a different scale and capturing specific frequency information.
	As illustrated in Fig.~\ref{fig:texture}, different frequency bands provide explicit information about the spatial distribution of details and textures at various scales. By replacing the deep-learning-based features with frequency-domain information obtained through Laplacian pyramid decomposition, we achieve comparable performance while eliminating the need for pre-trained networks (please refer to the analysis in section~\ref{sec:abla} for details). This reduces both the parameters and inference time.
	Specifically, we upsample different frequency bands to match the input resolution and apply learnable weights for band-by-band weighting. This allows the network to adaptively adjust the importance of different frequency bands for diverse videos.
	Next, we concatenate the fused frequency maps with the current frame $\bm x_i$, which serves as the multi-scale frequency prior, denoted by $\bm p_i$. 
	Inserting these frequency priors into the current model is straightforward: we replace all instances of $\bm x$ with $\bm p$ (except for the last residual connection which produces the HR video $\hat{\bm y}$). 
	
	\subsection{Bidirectional Flow-Guided Propagation Unit}
	Given $\bm x = \{\bm x_{i}\}_{i=0}^T$, the bidirectional flow-guided propagation unit computes two sequence of hidden states $\{\bm h^{\rightarrow}_{i}\}_{i=0}^{T}$ and $\{\bm h^{\leftarrow}_{i}\}_{i=0}^{T}$ to capture long-term spatiotemporal dependencies of previous and future frames. 
	It is worth emphasizing that we first fed the features with priors $\bm p$ into a ResNet with $N$ residual blocks for feature extraction, which serve as the original feature list $\bm{h}$.
	We take estimating the forward hidden states $\{\bm h^{\rightarrow}_{i}\}_{i=0}^{T}$ as an example.
	Initially, we estimate the optical flow between the current and previous frames~\cite{chan2021basicvsr}:
	\begin{align}
		\bm{f}_{i\rightarrow{i-1}} = \mathtt{flow}\left(\bm{x}_{i}, \bm{x}_{i-1}\right), \quad i \in \{1,2,\ldots, T\},
	\end{align}
	where $\mathtt{flow}(\cdot)$ denotes a state-of-the-art optical flow estimator~\cite{ranjan2017optical}. 
	$\bm{f}_{i\rightarrow{i-1}}$ is then used to align the hidden state $\bm h_{i-1}$ backward:
	\begin{align}\label{eq:calign}
		\bm h_{i-1\rightarrow i} = \mathtt{calign}(\bm h_{i-1}, \bm h_{i}, \bm f_{i \rightarrow i-1}), i \in \{1,2,\ldots, T\},
	\end{align}
	where $\mathtt{calign}(\cdot)$ denotes alignment based on motion compensation (discussed in the next subsection). Subsequently, the aligned previous hidden state $\bm h_{i-1\rightarrow i}$ and the current frame $\bm x_{i}$ are concatenated along the channel dimension and processed through a ResNet with $N$ residual blocks to compute $\bm h^{\rightarrow}_i$. The backward hidden state $\bm h^{\leftarrow}_i$ can be computed by reversing the input sequence and applying the aforementioned formulas. It also needs to be concatenated with the previously computed hidden state and fed into a ResNet with $N$ residual blocks. 
	As prior studies~\cite{chan2022basicvsr++} have demonstrated, multiple bidirectional information interaction refines features by propagating intermediate features forward and backward in an alternating manner over time. This allows information from different frames to be revisited for feature refinement. Hence, we employ a two-iteration bidirectional RNN propagation to compute bidirectional hidden states. Specifically, after obtaining forward and backward hidden states via the aforementioned operations, we repeat the process and substitute the original hidden states with the refined ones.
	To further enhance propagation robustness, we adopt second-order connections, which aggregate information from more spatiotemporal locations, improving the robustness and effectiveness of occluded and fine-detailed regions. More details are provided in the next section.
	The bidirectional flow-guided propagation unit allows the proposed BasicAVSR to incorporate long-term spatiotemporal context while being flow-aware. 
	\begin{figure*}[!t]\footnotesize
		\centering
		\setlength{\abovecaptionskip}{3pt} 
		\setlength{\belowcaptionskip}{0pt}
		\begin{tabular}{cccccc}
			\includegraphics[width=\linewidth]{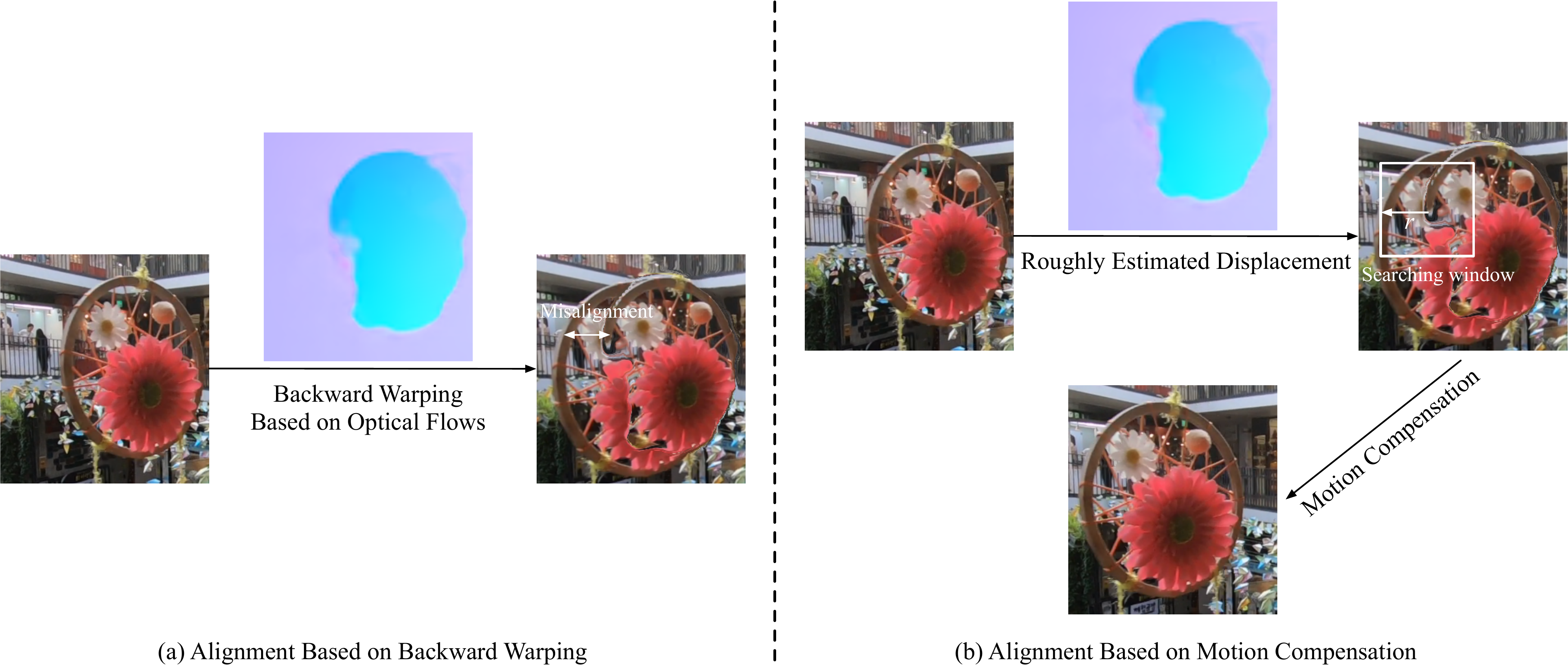}\\
		\end{tabular}
		%	\vspace{-1em}
		\caption{Comparison of traditional alignment and our proposed motion compensation. The displacement is roughly estimated based on the optical flow, and then a window of size $r$ is expanded in the adjacent frames with the roughly estimated pixel coordinates as the center to search for the pixel most similar to the source pixel to complete the motion compensation.
		}
		\label{fig:local}
	\end{figure*}

	\subsection{Second-Order Motion Compensation Unit} 
	The standard image/feature warping operation uses optical flow to align features or pixels from a neighboring frame to match the spatial location of the current frame. 
	This introduces two problems: (i) interpolation methods like bilinear or bicubic are generally employed for non-integer displacements, which are estimated without knowledge of the original downsampling kernel, so the smoothness prior inherent to most kernels yields overly-smooth, detail-losing results, (ii) any inaccuracy in the optical-flow estimation is directly encoded in the sampling grid and propagated to the output. 
	To address these limitations, we adopt a coarse-to-fine motion compensation strategy instead of traditional interpolation. Specifically, we first use the initially estimated optical flow to identify the approximate locations in adjacent frames that correspond to the current frame. Then, we conduct a precise window-based search around these locations to enhance alignment accuracy. Finally, we integrate advanced second-order alignment strategies from existing VSR techniques. This refines motion estimation and enhances the accuracy of aligning adjacent frames.

	Our alignment is inherently a two-step refinement process. We first perform the `coarse estimation' using a pre-trained optical flow network to predict the initial optical flows $\bm{f}$ between the current state $\bm{h}_{c}$ and the neighbouring state $\bm{h}_{n}$. This initial optical flow estimation roughly determines the spatial displacement range and establishes a local region of interest for the next step.
	Following the `coarse estimation', we perform a `fine-grained search-based matching' to obtain the precise sub-pixel offset.
	To help the network better capture spatial information and enhance feature discrimination, we utilize a position encoding network, which takes coordinates as inputs to model signals. This network serves as a prior for the `fine estimation', with the prior encoded as trainable weights within an MLP. MLPs are theoretically universal approximators capable of representing any function and frequency~\cite{hornik1989multilayer}. Especially showing strong learning ability for high-frequency content~\cite{mildenhall2021nerf}. 
	This matching process is formulated as an attention mechanism. Given the coordinate $p$, the aligned feature $\bm{h}_{n\rightarrow c}$ at the spatial position $p$ is aggregated by computing the similarity between the query patch computed by $\bm{h}_{c}$ at $p$ and the key patches computed by $\bm{h}_{n}$ in neighbors $p'$.
	The computation for `fine-grained search-based matching' is as follows:
	\begin{equation}
		\begin{aligned}
			\bm{h}_{n\rightarrow c}(p) &= \mathtt{Softmax}(\frac{\bm{q}\bm{k}^{\mathsf{T}}}{\sqrt{\gamma}})\bm{v},\\
			\bm{q} &=  \mathtt{MLP}(\bm{h}_{c}, \mathtt{SPE}(p)),  \\
			\bm{k} &=  \mathtt{MLP}(\bm{h}_{n}, \mathtt{SPE}(p')),  \\
			\bm{v} &=  \mathtt{MLP}(\bm{h}_{n}, \mathtt{SPE}(p')),
		\end{aligned}
	\end{equation}
	where $p'$ is the set of neighboring locations within a search window of size $r$ centered at the initial target position. This target position is coarsely estimated by displacing the coordinate $p$ according to the initial optical flows, which is the window center. 
	We employ sinusoidal positional encoding $\mathtt{SPE}(\cdot)$ as a pre-processing step to enhance the discriminability of coordinate-relevant inputs.
	$p'$ can be represented by the following formula:
	\begin{align}
		p' = \{p\,|\,\mathtt{Nearest}_{r}\left(p+\bm f(p)\right)\}
	\end{align}
	Based on the coarse optical flow estimation, we can confine the search to a small area around the estimated coordinates, which is more computationally efficient than conventional attention mechanisms. Specifically, the computational cost is reduced from the quadratic $O((HW)^2)$ of global attention to $O(r^2HW)$ for window-based attention. The window size $r$ can be adjusted according to different motion accuracy requirements. Generally, a larger $r$ is more robust to noisy motion estimation, while a smaller $r$ yields sharper results. 
	We use $\mathtt{calign}(\cdot)$ to denote the aforementioned two-stage alignment process.

	To further enhance the fidelity of the feature propagation, we adopt a second-order deformable alignment strategy inspired by BasicVSR++~\cite{chan2022basicvsr++}. This approach aligns adjacent two frames with the current frame, introducing a residual deformable convolution. This convolution learns residual offsets $\bm o$ and modulation masks $\bm m$ to locally correct the minor misalignments that previous steps might have missed, ensuring the final feature aggregation is accurately aligned.
	Finally, we use the $i$-th frame ($i\textgreater 1$) as an example to demonstrate second-order motion compensation, replacing Eq.~\eqref{eq:calign} with the following computation formula:
	\begin{equation}
		\small
		\!\!\!\!\! \!
		\begin{aligned}
			\bm h_{i-1\rightarrow i} &= \mathtt{calign}(\bm h_{i-1}, \bm h_{i}, \bm f_{i \rightarrow i-1}), \\
			\bm f_{i \rightarrow i-2} & = \bm f_{i \rightarrow i-1} + \mathtt{warp}(f_{i-1 \rightarrow i-2}, \bm f_{i \rightarrow i-1}), \\
			\bm h_{i-2\rightarrow i} &= \mathtt{calign}(\bm h_{i-2}, \bm h_{i}, \bm f_{i \rightarrow i-2}), \\
			\bm o_{i\rightarrow i-j} &= \bm f_{i \rightarrow i-j} + \mathtt{Conv}(\bm h_{i-j\rightarrow i}, \bm h_{i}, \bm h_{i-2\rightarrow i}), j=1,2, \\
			\bm m_{i\rightarrow i-j} &= \mathtt{Sigmoid}(\mathtt{Conv}(\bm h_{i-j\rightarrow i}, \bm h_{i}, \bm h_{i-2\rightarrow i})), j=1,2, \\
			\bm o_{i} &= \mathtt{Concat}(\bm o_{i\rightarrow i-1}, \bm o_{i\rightarrow i-2}), \\
			\bm m_{i} &= \mathtt{Concat}(\bm m_{i\rightarrow i-1}, \bm m_{i\rightarrow i-2}), \\
			\bm h'_{i} &= \mathtt{DCN}(\mathtt{Concat}(\bm h_{i-1}, \bm h_{i-2}); \bm o_{i}, \bm m_{i}), 
		\end{aligned}
	\end{equation}
	where $\mathtt{DCN}$ is the deformable convolution operation guided by learned offsets and modulation masks, $\mathtt{warp}(\cdot)$ denotes the standard backward warping operation using the bilinear kernel, $\mathtt{Concat}$ is the feature concatenation operation. For cases where $i\leq 1$, the hidden states and corresponding optical flows are initialized to $\bm{0}$.
	Finally, the aligned hidden state $\bm h'_{i}$ is used to update the original hidden state list. The calculation for the reverse RNN follows the same principle and will not be detailed here.
	This two-stage feature search combined with the final deformable refinement establishes a robust and high-precision temporal feature aggregation framework. 
	The features $\bm{g}$, after undergoing bidirectional propagation and precise alignment, are fed into the subsequent hyper-upsampling unit.

	\begin{figure*}[!t]\footnotesize
		\centering
		\setlength{\abovecaptionskip}{3pt} 
		\setlength{\belowcaptionskip}{0pt}
		%	\begin{tabular}{c}
			\includegraphics[width=0.75\linewidth]{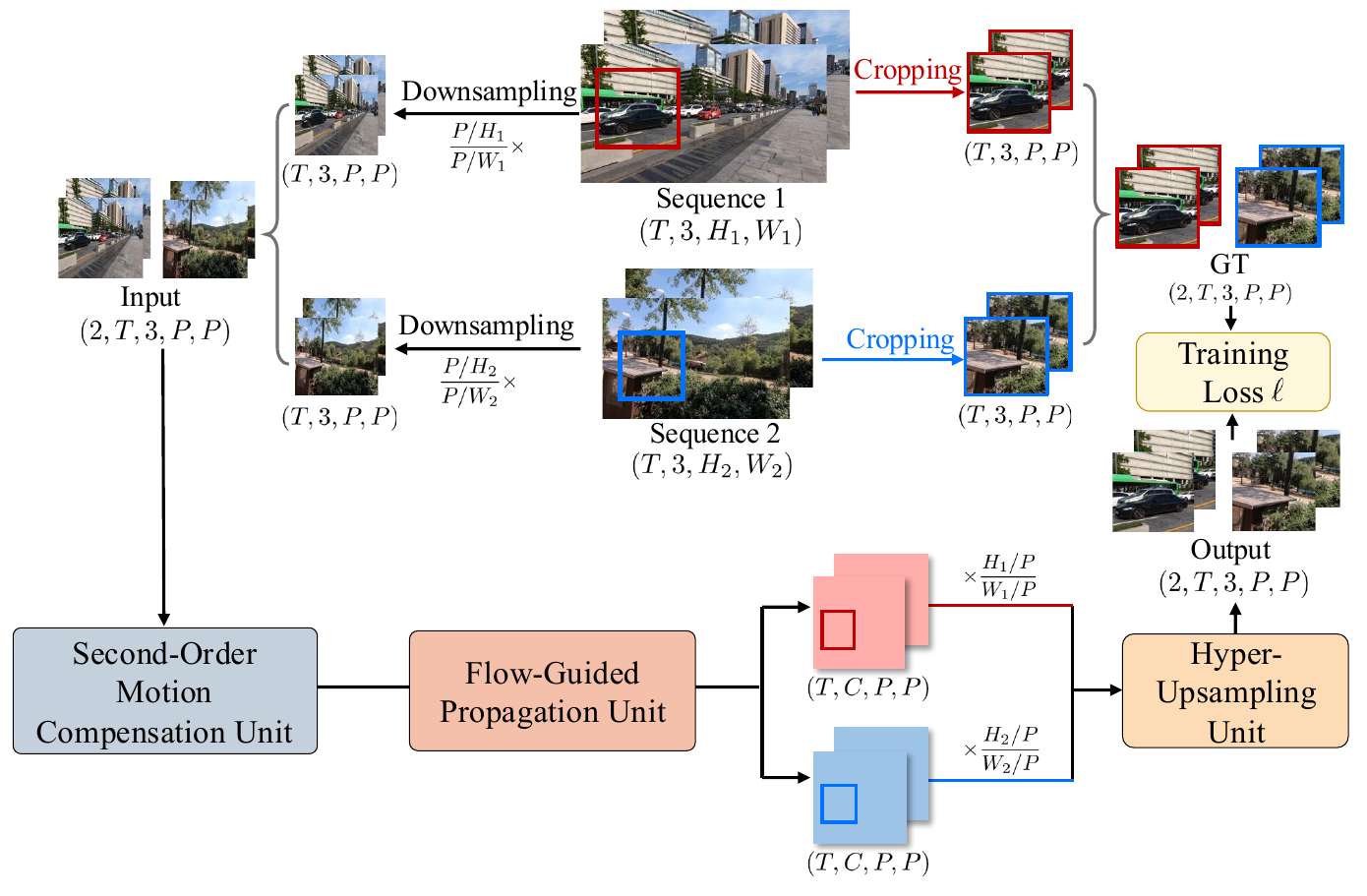}\\
			%	\end{tabular}
		\caption{Data pre-processing and training pipeline for BasicAVSR.}
		\label{fig:pipeline}
	\end{figure*} 
	\subsection{Hyper-Upsampling Unit}
	Inspired by the neural kriging upsampler~\cite{wang2023deep}, our hyper-upsampling unit consists of two branches: SR feature preparation and SR kernel prediction, as shown in Fig.~\ref{fig:framework}. For SR feature preparation, we pass the final aligned hidden state $\bm{g}_i$ through a ResNet with $N$ residual blocks to compute SR features.
	Next, we unfold a $K\times K$ spatial neighborhood of $C$-dimensional SR feature representations into $C\times K^2$ channels (\ie, the tensor generalization of $\mathtt{img2col}(\cdot)$ in image processing). Finally, we upsample the unfolded features to the target resolution using bilinear interpolation, resulting in $\bm{s}_{i}$.

	For SR kernel generation, we train a hyper-network, \ie, a
	multi-layer perceptron (MLP) with periodic activation functions~\cite{chen2023cascaded}, to predict the upsampling kernels $\bm w$.  
	% in which the four hidden dimensions are $16$, $16$, $16$, and $64$, respectively.
	Periodic activations have been shown to effectively address the spectral bias of MLPs, outperforming ReLU non-linearity~\cite{sitzmann2020implicit}. The inputs to the MLP are carefully selected to be scale-aware and content-independent. These include
	1) the scaling factors $(\alpha, \beta)$, 2) the relative coordinates between the LR and HR frames $(\delta_\alpha,\delta_\beta)$, and 3) the spatial indices $(k_1, k_2)$ of $\bm w$. The first two inputs have been used in other continuous representation methods~\cite{chen2021learning,lee2022local}.
	To enhance the discriminability of scale-relevant inputs, we employ sinusoidal positional encoding as a pre-processing step.
	It is noteworthy that our upsampling kernels $\bm w$ can be pre-computed and stored for various target resolutions, which accelerates inference time. 
	
	After obtaining $\bm w$, we perform Hadamard multiplication between $\bm w$ and $\bm{s}_{i}$, followed by a folding operation (\ie, the inverse of the unfolding operation).
	Finally, we employ a $1\times 1$ convolution to blend information across the channel dimension, followed by a $3\times 3$ convolution for channel adjustment, with LeakyReLU in between. The output from the last $3\times 3$ convolution layer is then added to the upsampled LR frame to produce the output $\hat{\bm y}_i$.
	
	\subsection{Architectural Variants: Adapting to Diverse Application Scenarios} \label{sec:arch_exten}
	In diverse VSR scenarios, bidirectional RNNs require processing and storing bidirectional hidden states for entire video sequences, which demands significant memory and makes such architectures unsuitable for online applications. 
	To meet online requirements, we extend the BasicAVSR framework to a unidirectional RNN variant by removing the backward RNN and keeping only the forward RNN. For scenarios allowing slight latency, we integrate a lookahead strategy that leverages limited future frames, drawing on the strategy used in our previous work ST-AVSR~\cite{shang2024arbitrary}. Specifically, in the ST-AVSR, the alignment strategy and priors are replaced with the proposed image priors and enhanced motion compensation, respectively.
	%Fig.~\ref{fig:extend} illustrates the overall architecture of these three variants: unidirectional RNN, unidirectional RNN with lookahead, and bidirectional RNN. 
	%
	In a unidirectional RNN, only the hidden state of the previous frame is stored, and it is overwritten by the new hidden state after computing the current frame, enabling online output. For the unidirectional RNN with lookahead, in addition to the hidden state of the previous frame, it also requires the hidden states of the next $L$ frames. Thus, the total number of stored hidden states is $L+1$, resulting in an output delay of $L$ frames. A bidirectional RNN needs to store the bidirectional hidden states of the entire video sequence, with the total number of stored hidden states being $2T$, making it only suitable for offline processing. All variants achieve state-of-the-art performance (please see Sec~\ref{sec:extension} for detailed analysis).

	\begin{table*}[!t]
		\centering
		\caption{Quantitative comparison with state-of-the-art methods on the REDS validation set (PSNR$\uparrow$ / SSIM$\uparrow$ / LPIPS$\downarrow$). The best results are highlighted in boldface. }
		\label{table:reds}
		\renewcommand{\arraystretch}{1.2}
		%	\resizebox{\columnwidth}{!}{ GaussianSR~\cite{hu2025gaussiansr} and ContinuousSR~\cite{peng2025pixel}. BF-STVSR~\cite{kim2025bf}.
			\begin{tabular}{cc|c|c|c|c|c}
				\multicolumn{2}{c|}{Method} & \multicolumn{5}{c}{Scale} \\  
				\hline
				\multirow{2}{*}{Backbone} & Upsampling  & \multirow{2}{*}{$\times 2$} & \multirow{2}{*}{$\times 3$} & \multirow{2}{*}{$\times 4$} & \multirow{2}{*}{$\times 6$} & \multirow{2}{*}{$\times 8$}  \\ 
				&  Unit & & & & & \\ \hline   %\hline 
				\multicolumn{2}{c|}{Bicubic} &  {\fontsize{8}{10}\selectfont 31.51/0.911/0.165}   & {\fontsize{8}{10}\selectfont 26.82/0.788/0.377}  & {\fontsize{8}{10}\selectfont  24.92/0.713/0.484}  & {\fontsize{8}{10}\selectfont  22.89/0.622/0.631}  &  {\fontsize{8}{10}\selectfont 21.69/0.574/0.699}  \\ 
				\multicolumn{2}{c|}{EDVR~\cite{wang2019edvr}} &  {\fontsize{8}{10}\selectfont 36.03/0.961/0.072}   & {\fontsize{8}{10}\selectfont 32.59/0.904/0.108}  & {\fontsize{8}{10}\selectfont  30.24/0.853/0.202}  & {\fontsize{8}{10}\selectfont  27.02/0.733/0.349}  &  {\fontsize{8}{10}\selectfont 25.38/0.678/0.411}  \\
				\multicolumn{2}{c|}{ArbSR~\cite{wang2021learning}} & {\fontsize{8}{10}\selectfont 34.48/0.942/0.096} & {\fontsize{8}{10}\selectfont 30.51/0.862/0.200} &  {\fontsize{8}{10}\selectfont 28.38/0.799/0.295} & {\fontsize{8}{10}\selectfont 26.32/0.710/0.428} & {\fontsize{8}{10}\selectfont 25.08/0.641/0.492}  \\
				\multicolumn{2}{c|}{EQSR~\cite{wang2023deep}}   & {\fontsize{8}{10}\selectfont 34.71/0.943/0.082}  & {\fontsize{8}{10}\selectfont 30.71/0.867/0.194}    & {\fontsize{8}{10}\selectfont 28.75/0.804/0.283}        & {\fontsize{8}{10}\selectfont 26.53/0.718/0.391}      & {\fontsize{8}{10}\selectfont 25.23/0.645/0.459}    \\ \hline 
				\multirow{5}{*}{RDN~\cite{zhang2018residual}} & LTE~\cite{lee2022local} & {\fontsize{8}{10}\selectfont 34.63/0.942/0.093}  & {\fontsize{8}{10}\selectfont 30.64/0.865/0.204}     & {\fontsize{8}{10}\selectfont 28.65/0.801/0.289}     & {\fontsize{8}{10}\selectfont 26.46/0.714/0.410}    & {\fontsize{8}{10}\selectfont 25.15/0.660/0.488} \\ 
				& CLIT~\cite{chen2023cascaded} & {\fontsize{8}{10}\selectfont 34.63/0.942/0.092}   & {\fontsize{8}{10}\selectfont 30.63/0.865/0.204}       & {\fontsize{8}{10}\selectfont 28.63/0.801/0.290}           & {\fontsize{8}{10}\selectfont  26.43/0.714/0.400}    & {\fontsize{8}{10}\selectfont 25.14/0.661/0.467}\\ 
				& OPE~\cite{song2023ope} & {\fontsize{8}{10}\selectfont 34.05/0.939/0.082}   & {\fontsize{8}{10}\selectfont 30.52/0.864/0.199}    & {\fontsize{8}{10}\selectfont 28.63/0.800/0.293}   & {\fontsize{8}{10}\selectfont 26.37/0.711/0.421}    & {\fontsize{8}{10}\selectfont 25.04/0.655/0.504} \\ 
				& GaussianSR~\cite{hu2025gaussiansr} & {\fontsize{8}{10}\selectfont 34.25/0.940/0.091}   & {\fontsize{8}{10}\selectfont 30.56/0.866/0.201}    & {\fontsize{8}{10}\selectfont 28.64/0.800/0.291}   & {\fontsize{8}{10}\selectfont 26.40/0.712/0.419}    & {\fontsize{8}{10}\selectfont 25.08/0.657/0.501} \\ 
				& ContinuousSR~\cite{peng2025pixel} & {\fontsize{8}{10}\selectfont ---/---/---}   & {\fontsize{8}{10}\selectfont 30.65/0.866/0.198}    & {\fontsize{8}{10}\selectfont 28.67/0.801/0.289}   & {\fontsize{8}{10}\selectfont 26.49/0.715/0.402}    & {\fontsize{8}{10}\selectfont 25.14/0.662/0.470} \\ 
				\hline 
				\multirow{5}{*}{SwinIR~\cite{liang2021swinir}} & LTE~\cite{lee2022local} & {\fontsize{8}{10}\selectfont 34.73/0.943/0.091}      & {\fontsize{8}{10}\selectfont 30.73/0.866/0.200}     & {\fontsize{8}{10}\selectfont 28.75/0.804/0.284}      & {\fontsize{8}{10}\selectfont 26.56/0.718/0.403}         & {\fontsize{8}{10}\selectfont 25.24/0.669/0.480} \\ 
				& CLIT~\cite{chen2023cascaded} & {\fontsize{8}{10}\selectfont 34.63/0.942/0.093}       & {\fontsize{8}{10}\selectfont 30.64/0.865/0.205}      & {\fontsize{8}{10}\selectfont 28.64/0.802/0.291}     & {\fontsize{8}{10}\selectfont 26.45/0.715/0.400}     & {\fontsize{8}{10}\selectfont 25.15/0.662/0.466} \\ 
				& OPE~\cite{song2023ope} & {\fontsize{8}{10}\selectfont  33.39/0.935/0.081}     & {\fontsize{8}{10}\selectfont 29.40/0.820/0.217}   & {\fontsize{8}{10}\selectfont 28.49/0.785/0.292}          & {\fontsize{8}{10}\selectfont 26.30/0.698/0.398}    & {\fontsize{8}{10}\selectfont 25.01/0.648/0.487}  \\
				& GaussianSR~\cite{hu2025gaussiansr} & {\fontsize{8}{10}\selectfont 34.31/0.941/0.089}   & {\fontsize{8}{10}\selectfont 30.60/0.867/0.199}    & {\fontsize{8}{10}\selectfont 28.69/0.802/0.290}   & {\fontsize{8}{10}\selectfont 26.42/0.713/0.416}    & {\fontsize{8}{10}\selectfont 25.08/0.659/0.498} \\ 
				& ContinuousSR~\cite{peng2025pixel} & {\fontsize{8}{10}\selectfont ---/---/---}   & {\fontsize{8}{10}\selectfont 30.75/0.868/0.197}    & {\fontsize{8}{10}\selectfont 28.68/0.805/0.287}   & {\fontsize{8}{10}\selectfont 26.58/0.720/0.401}    & {\fontsize{8}{10}\selectfont 25.26/0.670/0.467} \\ 
				\hline 
				
				\multicolumn{2}{c|}{VideoINR~\cite{chen2022videoinr}}    &  {\fontsize{8}{10}\selectfont 31.59/0.900/0.144}  &  {\fontsize{8}{10}\selectfont 30.04/0.852/0.197}    & {\fontsize{8}{10}\selectfont 28.13/0.791/0.263}   & {\fontsize{8}{10}\selectfont  25.27/0.687/0.374} & {\fontsize{8}{10}\selectfont 23.46/0.619/0.470}  \\ 
				\multicolumn{2}{c|}{MoTIF~\cite{chen2023motif}}      & {\fontsize{8}{10}\selectfont 31.03/0.898/0.100}      & {\fontsize{8}{10}\selectfont 30.44/0.862/0.186}  & {\fontsize{8}{10}\selectfont 28.77/0.807/0.260}    & {\fontsize{8}{10}\selectfont 25.63/0.698/0.369}    & {\fontsize{8}{10}\selectfont 25.12/0.664/0.467}     \\
				\multicolumn{2}{c|}{BF-STVSR~\cite{kim2025bf}}  & {\fontsize{8}{10}\selectfont 32.06/0.908/0.092}      & {\fontsize{8}{10}\selectfont 31.38/0.877/0.146}  & {\fontsize{8}{10}\selectfont 29.29/0.837/0.200}    & {\fontsize{8}{10}\selectfont 25.98/0.718/0.321}    & {\fontsize{8}{10}\selectfont 25.42/0.670/0.459}     \\
				\multicolumn{2}{c|}{SAVSR~\cite{li2024savsr}}     & {\fontsize{8}{10}\selectfont 35.66/0.955/0.046}  & {\fontsize{8}{10}\selectfont 32.19/0.918/0.100}     & {\fontsize{8}{10}\selectfont 30.61/0.872/0.138}     & {\fontsize{8}{10}\selectfont 27.03/0.791/0.250}    & {\fontsize{8}{10}\selectfont 25.59/0.716/0.312}      \\
				\multicolumn{2}{c|}{ST-AVSR~\cite{shang2024arbitrary}}     & {\fontsize{8}{10}\selectfont 36.91/0.969/0.041}    & {\fontsize{8}{10}\selectfont  33.41/0.937/0.066}       & {\fontsize{8}{10}\selectfont  31.03/0.897/0.114}      & {\fontsize{8}{10}\selectfont 27.89/0.812/0.222}    & {\fontsize{8}{10}\selectfont 26.04/0.746/0.298}     \\
				\hline 
				\multicolumn{2}{c|}{BasicAVSR (Ours)}     & {\fontsize{8}{10}\selectfont \textbf{37.40}/\textbf{0.972}/\textbf{0.040}}    & {\fontsize{8}{10}\selectfont  \textbf{34.82}/\textbf{0.954}/\textbf{0.050}}       & {\fontsize{8}{10}\selectfont  \textbf{32.74}/\textbf{0.931}/\textbf{0.081}}      & {\fontsize{8}{10}\selectfont \textbf{29.45}/\textbf{0.864}/\textbf{0.167}}    & {\fontsize{8}{10}\selectfont \textbf{27.33}/\textbf{0.798}/\textbf{0.247}}     \\ 
				
			\end{tabular}
			%}
	\end{table*}
	
	\begin{figure*}[!t]\footnotesize
		\centering
		\setlength{\abovecaptionskip}{3pt} 
		\setlength{\belowcaptionskip}{0pt}
		\begin{tabular}{cccccc}
			\includegraphics[width=0.85\linewidth]{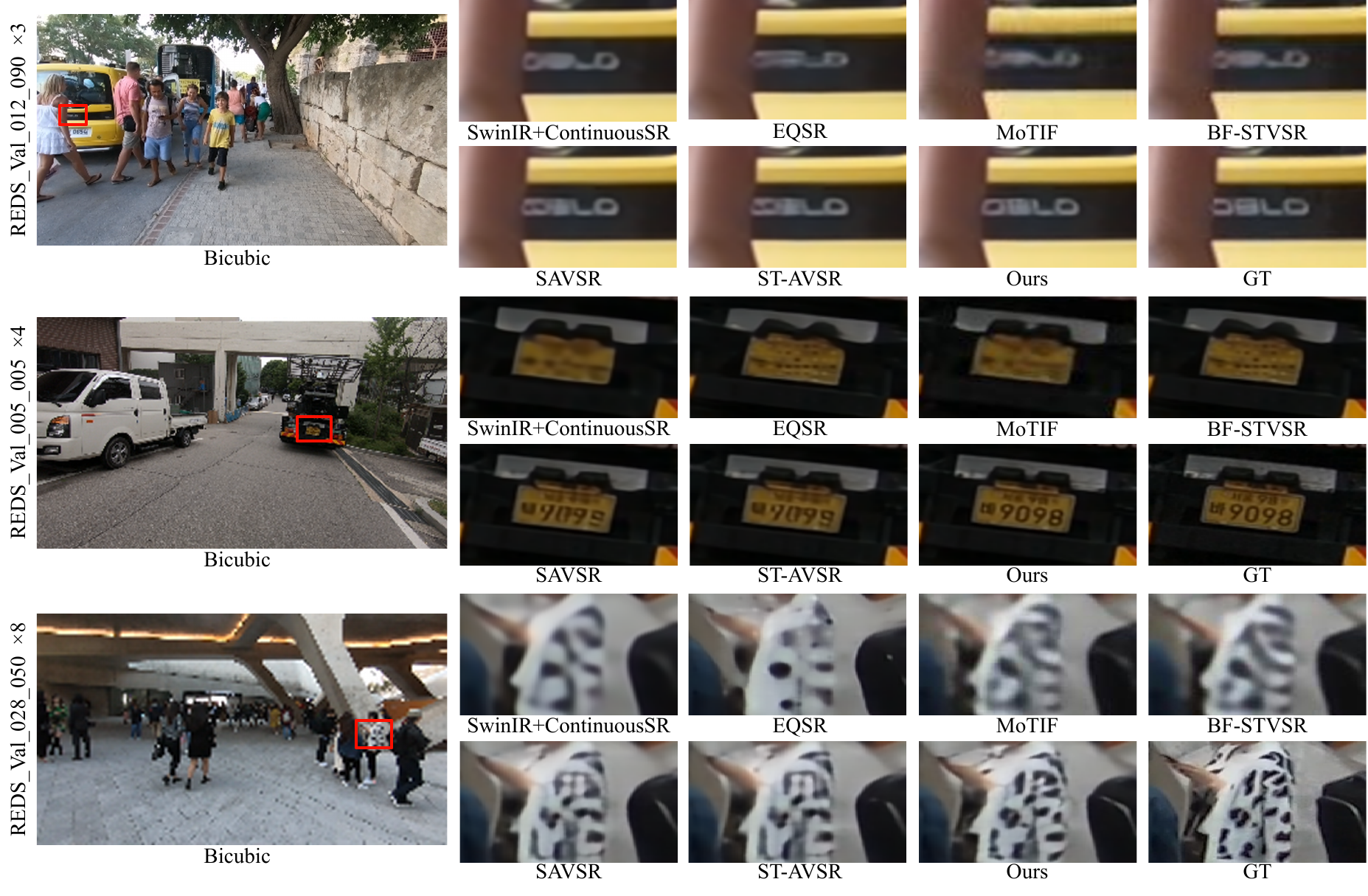}\\
		\end{tabular}
		\caption{
			Visual comparison of different AVSR methods on the REDS dataset. Zoom in for better distortion visibility.
		}
		\label{fig:reds}
	\end{figure*}
	
	\begin{table*}[!t]
		\centering
		\caption{Quantitative comparison with state-of-the-art methods for {AVSR} on the Vid4 dataset (PSNR$\uparrow$ / SSIM$\uparrow$ / LPIPS$\downarrow$). }
		\label{table:vid4}
		\renewcommand{\arraystretch}{1.2}
		%	\resizebox{\columnwidth}{!}{
			\begin{tabular}{cc|c|c|c|c}
				\multicolumn{2}{c|}{Method} & \multicolumn{4}{c}{Scale}  \\ 
				%			\cline{1-5}
				\hline
				\multirow{2}{*}{Backbone} & Upsampling   & \multirow{2}{*}{$\times \frac{2.5}{3.5}$} & \multirow{2}{*}{$\times \frac{4}{4}$} & \multirow{2}{*}{$\times \frac{7.2}{6}$}  & \multirow{2}{*}{$\times \frac{6.4}{9}$} \\ 
				&  Unit & & & &  \\ \hline %\hline 
				\multicolumn{2}{c|}{Bicubic} & {\fontsize{8}{10}\selectfont 23.00/0.728/0.396} & {\fontsize{8}{10}\selectfont 20.96/0.617/0.498} & {\fontsize{8}{10}\selectfont 18.73/0.463/0.691} &  {\fontsize{8}{10}\selectfont 18.15/0.430/0.732} \\ 
				\multicolumn{2}{c|}{ArbSR~\cite{wang2021learning}} & {\fontsize{8}{10}\selectfont 25.86/0.815/0.224} & {\fontsize{8}{10}\selectfont 24.01/0.721/0.313} &  {\fontsize{8}{10}\selectfont 21.23/0.540/0.478} & {\fontsize{8}{10}\selectfont  20.34/0.515/0.498}   \\
				\multicolumn{2}{c|}{EQSR~\cite{wang2023deep}}   & {\fontsize{8}{10}\selectfont  26.24/0.826/0.210}  & {\fontsize{8}{10}\selectfont 24.16/0.730/0.300}    & {\fontsize{8}{10}\selectfont {21.72}/0.573/0.443}        & {\fontsize{8}{10}\selectfont  20.81/0.528/0.472}      \\ \hline 
				\multirow{5}{*}{RDN~\cite{zhang2018residual}} & LTE~\cite{lee2022local} & {\fontsize{8}{10}\selectfont  25.98/0.818/0.226}  & {\fontsize{8}{10}\selectfont  24.03/0.722/0.312}     & {\fontsize{8}{10}\selectfont  21.64/0.565/0.455}     & {\fontsize{8}{10}\selectfont 20.60/0.522/0.480}    \\ 
				& CLIT~\cite{chen2023cascaded}  & {\fontsize{8}{10}\selectfont  25.83/0.815/0.223}   & {\fontsize{8}{10}\selectfont  23.94/0.721/0.312}      & {\fontsize{8}{10}\selectfont  21.62/0.563/0.458}           & {\fontsize{8}{10}\selectfont  20.57/0.520/0.491}   \\ 
				& OPE~\cite{song2023ope}  & {\fontsize{8}{10}\selectfont 25.77/0.818/0.217}  & {\fontsize{8}{10}\selectfont 23.98/0.719/0.317}    & {\fontsize{8}{10}\selectfont 21.60/0.559/0.483}   & {\fontsize{8}{10}\selectfont 20.55/0.528/0.495}   \\ 
				& GaussianSR~\cite{hu2025gaussiansr} & {\fontsize{8}{10}\selectfont 25.81/0.817/0.222}  & {\fontsize{8}{10}\selectfont 23.99/0.720/0.313}    & {\fontsize{8}{10}\selectfont 21.61/0.560/0.460}   & {\fontsize{8}{10}\selectfont 20.56/0.520/0.484}   \\
				& ContinuousSR~\cite{peng2025pixel}  & {\fontsize{8}{10}\selectfont 25.94/0.820/0.216}  & {\fontsize{8}{10}\selectfont 24.08/0.725/0.310}    & {\fontsize{8}{10}\selectfont 21.66/0.568/0.453}   & {\fontsize{8}{10}\selectfont 20.69/0.525/0.473}   \\ \hline 
				\multirow{5}{*}{SwinIR~\cite{liang2021swinir}} & LTE~\cite{lee2022local}    & {\fontsize{8}{10}\selectfont 26.43/0.826/0.217}   & {\fontsize{8}{10}\selectfont  24.09/0.727/0.305}     & {\fontsize{8}{10}\selectfont {21.72}/0.570/0.448}      & {\fontsize{8}{10}\selectfont 20.70/0.524/0.475}         \\ 
				& CLIT~\cite{chen2023cascaded} & {\fontsize{8}{10}\selectfont  25.89/0.818/0.224}   & {\fontsize{8}{10}\selectfont  24.00/0.724/0.314}         & {\fontsize{8}{10}\selectfont 21.65/0.565/0.457}     & {\fontsize{8}{10}\selectfont  20.69/0.522/0.479}  \\ 
				& OPE~\cite{song2023ope} & {\fontsize{8}{10}\selectfont 25.55/0.801/0.221} & {\fontsize{8}{10}\selectfont  23.93/0.711/0.320}     & {\fontsize{8}{10}\selectfont 21.58/0.521/0.471}          & {\fontsize{8}{10}\selectfont 20.65/0.520/0.492}    \\ 
				& GaussianSR~\cite{hu2025gaussiansr}   & {\fontsize{8}{10}\selectfont 25.92/0.820/0.220}  & {\fontsize{8}{10}\selectfont 24.01/0.722/0.311}    & {\fontsize{8}{10}\selectfont 21.63/0.563/0.455}   & {\fontsize{8}{10}\selectfont 20.66/0.523/0.480}   \\
				& ContinuousSR~\cite{peng2025pixel}  & {\fontsize{8}{10}\selectfont 26.54/0.830/0.210}  & {\fontsize{8}{10}\selectfont 24.16/0.729/0.301}    & {\fontsize{8}{10}\selectfont 21.76/0.573/0.444}   & {\fontsize{8}{10}\selectfont 20.80/0.539/0.469}   \\ 
				\hline 
				\multicolumn{2}{c|}{VideoINR~\cite{chen2022videoinr}}   &  {\fontsize{8}{10}\selectfont 23.02/0.715/0.203}  &  {\fontsize{8}{10}\selectfont 24.34/0.741/0.249}    & {\fontsize{8}{10}\selectfont 20.80/0.536/0.431}   & {\fontsize{8}{10}\selectfont  20.43/0.511/0.453}  \\ 
				\multicolumn{2}{c|}{MoTIF~\cite{chen2023motif}}   & {\fontsize{8}{10}\selectfont 23.55/0.734/0.209}    & {\fontsize{8}{10}\selectfont 24.52/0.746/0.261}      & {\fontsize{8}{10}\selectfont 20.94/0.546/0.426}    & {\fontsize{8}{10}\selectfont 20.48/0.518/0.450}     \\ 
				\multicolumn{2}{c|}{BF-STVSR~\cite{kim2025bf}}  &  {\fontsize{8}{10}\selectfont 24.12/0.745/0.166}  &  {\fontsize{8}{10}\selectfont 24.90/0.784/0.222}    & {\fontsize{8}{10}\selectfont 21.13/0.579/0.423}   & {\fontsize{8}{10}\selectfont  20.59/0.537/0.447}  \\ 
				\multicolumn{2}{c|}{SAVSR~\cite{li2024savsr}}   &  {\fontsize{8}{10}\selectfont 27.82/0.875/0.088}  &  {\fontsize{8}{10}\selectfont 25.97/0.835/0.154}    & {\fontsize{8}{10}\selectfont 21.42/0.645/0.359}   & {\fontsize{8}{10}\selectfont  20.73/0.588/0.393}  \\ 
				\multicolumn{2}{c|}{ST-AVSR~\cite{shang2024arbitrary}}    & {\fontsize{8}{10}\selectfont 29.09/0.913/0.069}    & {\fontsize{8}{10}\selectfont 26.16/0.852/0.127}      & {\fontsize{8}{10}\selectfont 21.60/0.668/0.306}    & {\fontsize{8}{10}\selectfont 20.64/0.609/0.357}     \\  %&  {\fontsize{8}{10}\selectfont 34.28/0.970/0.064}
				
				\hline
				\multicolumn{2}{c|}{BasicAVSR (Ours)}    & {\fontsize{8}{10}\selectfont  \textbf{30.32}/\textbf{0.934}/\textbf{0.058}}     & {\fontsize{8}{10}\selectfont \textbf{27.96}/\textbf{0.893}/\textbf{0.096}}       & {\fontsize{8}{10}\selectfont \textbf{22.40}/\textbf{0.724}/\textbf{0.257}}      & {\fontsize{8}{10}\selectfont \textbf{21.34}/\textbf{0.666}/\textbf{0.307}}     
				\\ 
			\end{tabular}
			%}
	\end{table*}
	
	\begin{figure*}[!t]\footnotesize
		\centering
		\setlength{\abovecaptionskip}{3pt} 
		\setlength{\belowcaptionskip}{0pt}
		\begin{tabular}{l}
			\includegraphics[width=0.95\linewidth]{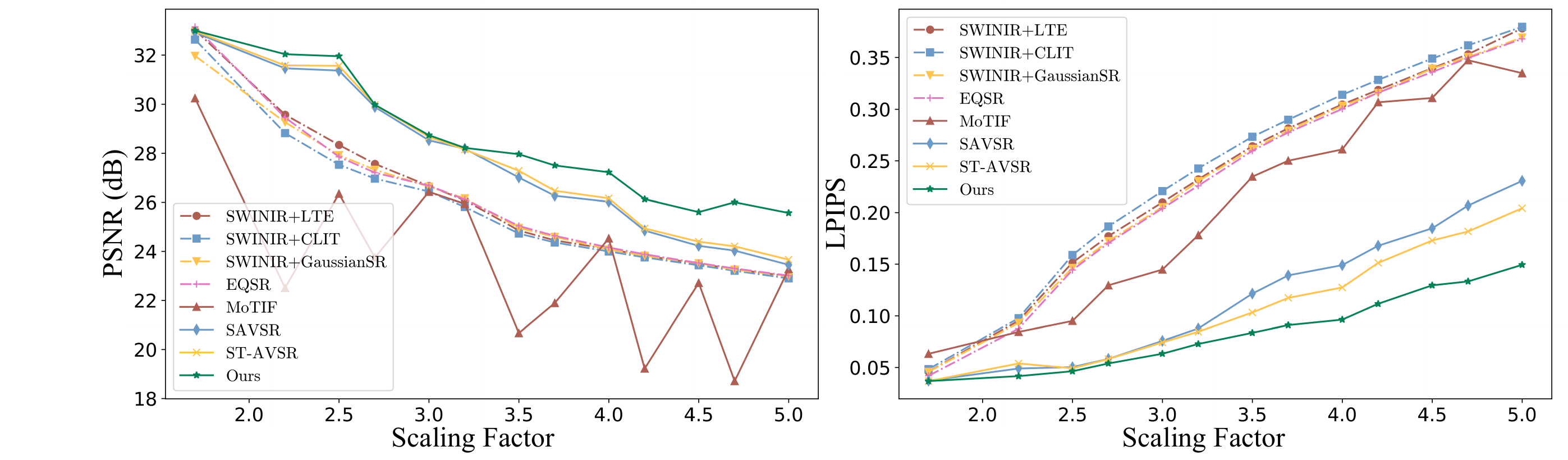}\\
		\end{tabular}
		%	\vspace{-1em}
		\caption{
			PSNR and LPIPS variations for different scaling factors on Vid4.
		}
		\label{fig:plot}
	\end{figure*}
	\section{Experiments}
	In this section, we first describe the experimental setups and then compare the proposed BasicAVSR against state-of-the-art AISR and AVSR methods, followed by a series of ablation studies to demonstrate the rationality of the key design updates in BasicAVSR. 
	Furthermore, we compare the performance of all propagation variants derived from BasicAVSR, and the results verify that each variant remains effective.
	\subsection{Experimental Setups}
	\subsubsection{Datasets}
	BasicAVSR is trained on the REDS dataset~\cite{nah2019ntire}, which comprises $240$ videos of resolution $720\times1,280$ captured by GoPro. Each video consists of $100$ HR frames. 
	Following the settings in~\cite{chen2022videoinr,chen2023cascaded,chen2023motif}, we generate LR frames using the bicubic degradation model, with randomly sampled scaling factors $(\alpha, \beta)$ from a uniform distribution $\mathcal{U}[1, 4]$. We test BasicAVSR on the validation set of REDS comprising $30$ videos, and the Vid4 dataset~\cite{liu2013bayesian} containing $4$ videos. To evaluate the generalization of our method to unseen degradation models, we applied a video random degradation pipeline~\cite{chan2022investigating} to the test set of GoPro~\cite{Nah_2017_CVPR}, incorporating noise and video compression to synthesize unseen degradations for validation. Additionally, we also use real-world and online collected data to verify the generalization of our method.
	\subsubsection{Data Pre-processing for Training}
	To enable mini-batch training with varying LR/HR resolutions, we adapt the pre-processing method used for AISR in EQSR~\cite{wang2023deep} to AVSR.  %$\alpha P\times \beta P\times T$
	Specifically, from an HR video patch of size $H\times W\times T$, we generate the input LR video patch by \textit{resizing} it to $P\times P\times T$. We next \textit{crop} a set of ground-truth patches of size  $P\times P\times T$  from the same HR patch. The respective relative coordinates $(\delta_\alpha, \delta_\beta)$ are recorded for use in the hyper-upsampling unit to differentiate between different ground-truth patches for the same input (see the data pre-processing pipeline in Fig.~\ref{fig:pipeline}). Data augmentation techniques include random rotation (by $90^\circ$, $180^\circ$, or $270^\circ$) and random horizontal and vertical flipping.

	\subsubsection{Implementation Details}
	BasicAVSR is end-to-end optimized for $300$K iterations. Adam~\cite{kingma2014adam} is chosen as the optimizer, with an initial learning rate $2\times$10$^{-4}$ 
	that is gradually lowered to 1$\times$10$^{-6}$ by cosine annealing~\cite{loshchilov2016sgdr}. 
	We set the input patch size to $P=80$, the sequence length to $T=15$,  the number of ResBlocks to $N=5$, the searching window size to $r=2$, the unfolding neighborhood to $K=3$, and the SR feature dimension to $C=64$, respectively.
	%, the sliding window size to $L=2$
	The hidden dimensions of the MLP in the hyper-upsampling unit are $16$, $16$, $16$, and $64$, respectively.
	The parameters of SPYNet~\cite{ranjan2017optical} as the optical flow estimator are frozen during training.
	We use the Charbonnier loss~\cite{lai2017deep}:
	\begin{equation}
		\ell(\hat{\bm y}, \bm y) = \frac{1}{(T+1)\vert\mathcal{Z}\vert} \sum^{T}_{i=0}\sum_{z\in\mathcal{Z}} \sqrt{ (\hat{\bm y}_{i}(z) - \bm{y}_{i}(z))^2 + \epsilon},
	\end{equation}
	where $z\in\mathcal{Z}$ denotes the spatial index, and $\vert\mathcal{Z}\vert$ is the number of all spatial indices. $\bm y $ indicates the ground-truth HR video sequence and $\epsilon$ is a smoothing parameter set to 1$\times$10$^{-9}$ in our experiments.

	\begin{figure*}[!t]\footnotesize
		\centering
		\setlength{\abovecaptionskip}{3pt} 
		\setlength{\belowcaptionskip}{0pt}
		\begin{tabular}{cccccc}
			\includegraphics[width=0.85\linewidth]{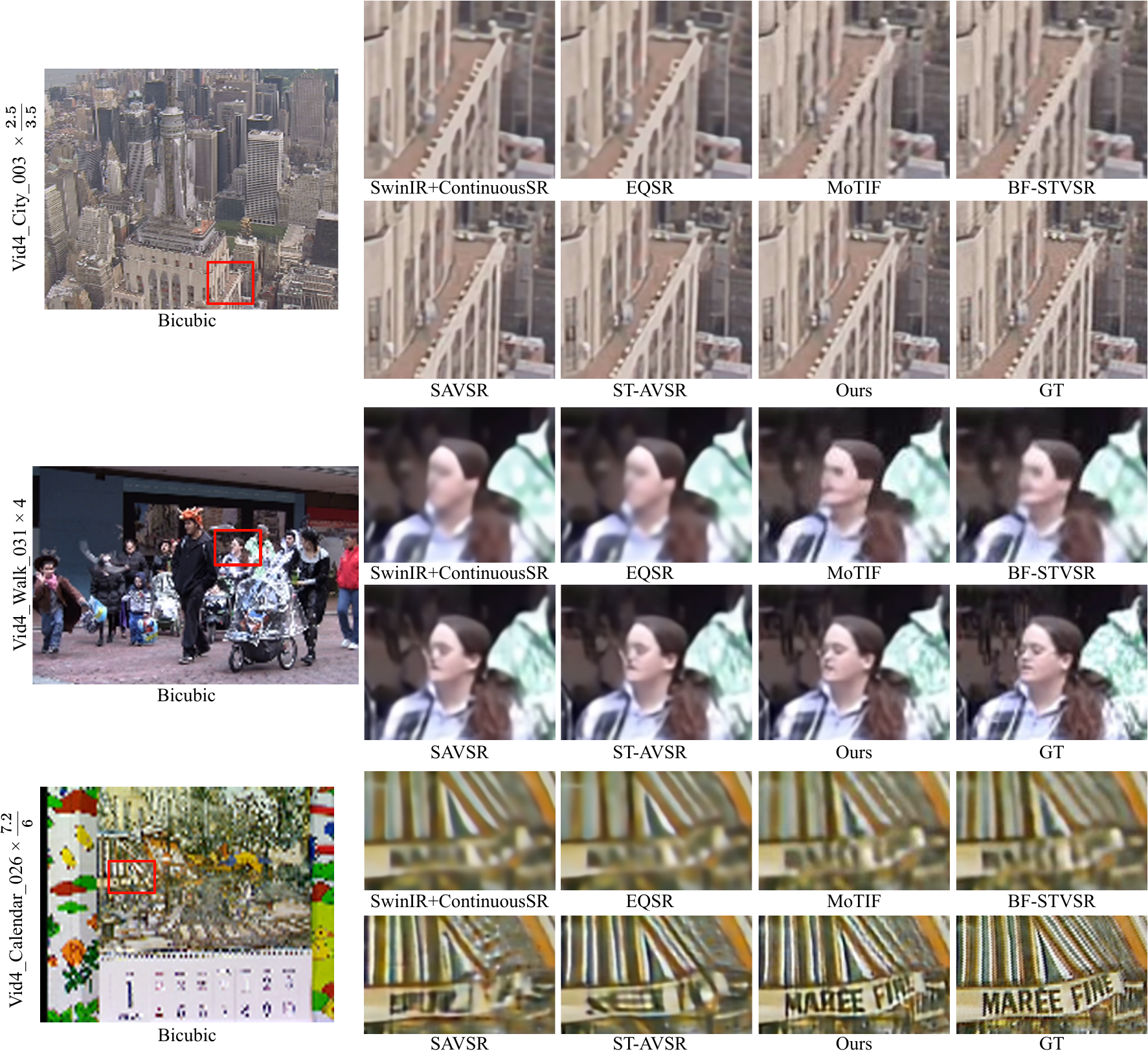}\\
		\end{tabular}
		%	\vspace{-1em}
		\caption{
			Visual comparison of different AVSR methods on Vid4.
			% Please zoom in for better view.
		}
		\label{fig:vid4}
	\end{figure*}
	\subsection{Comparison with State-of-the-art Methods}
	We compare BasicAVSR with state-of-the-art AISR and AVSR methods.
	For AISR, we choose methods from three categories: 1) learnable adaptive filter-based upsampling, including ArbSR~\cite{wang2021learning} and EQSR~\cite{wang2023deep}, 2) implicit neural representation-based upsampling, including LTE~\cite{lee2022local}, CLIT~\cite{chen2023cascaded}, OPE~\cite{song2023ope}, and 3) Gaussian splatting-based upsampling, including GaussianSR~\cite{hu2025gaussiansr} and ContinuousSR~\cite{peng2025pixel}. 
	For AVSR, we compare with VideoINR~\cite{chen2022videoinr}, MoTIF~\cite{chen2023motif}, SAVSR~\cite{li2024savsr}, ST-AVSR~\cite{shang2024arbitrary}, and BF-STVSR~\cite{kim2025bf}.
	%
	%Additionally, we include EDVR~\cite{wang2019edvr}, a state-of-the-art VSR method for integer scaling factors.   
	%
	All competing methods have been finetuned on the REDS dataset for a fair comparison, and we evaluate their generalization ability on Vid4~\cite{liu2013bayesian} and further assess robustness to unknown degradations by testing on GoPro~\cite{Nah_2017_CVPR} with a random degradation pipeline~\cite{chan2022investigating} and real-world data.
	More video results are available at the link \href{https://1drv.ms/f/s!AnS8oR69EOSMiVW-tj5ZHs5qYrFI?e=1syN9v}{\textbf{\emph{Video Results}}}.
	
	\begin{table*}[!t]
		\centering
		\caption{Comparison of the generalization on GoPro for $\times$4 SR under unseen degradations, along with an efficiency comparison in terms of parameters, complexity, and inference time.}
		\label{table:gen_gopro}
		\renewcommand{\arraystretch}{1.2}
		%	\resizebox{\columnwidth}{!}{
			\begin{tabular}{cc|c|c|c|c}
				\multicolumn{2}{c|}{Method} & \multirow{2}{*}{PSNR$\uparrow$ / SSIM$\uparrow$ / LPIPS$\downarrow$} &  \multirow{2}{*}{Parameters (M)} &  \multirow{2}{*}{Complexity (GFLOPs)}  &  \multirow{2}{*}{Inference Time (s)}  \\ 
				\cline{1-2}
				
				Backbone & Upsampling Unit   &  &  &     &  \\   \hline %\hline 
				\multicolumn{2}{c|}{Bicubic} & {\fontsize{8}{10}\selectfont 23.63/0.711/0.416} & {---} & {---} &  {---} \\ 
				\multicolumn{2}{c|}{ArbSR~\cite{wang2021learning}} & {\fontsize{8}{10}\selectfont 27.43/0.798/0.239} & 16.6  &  887.3 & 0.651  \\
				\multicolumn{2}{c|}{EQSR~\cite{wang2023deep}}   & {\fontsize{8}{10}\selectfont  28.00/0.815/0.228}  &  11.6  &  1743.2        & 0.921      \\ \hline 
				\multirow{5}{*}{RDN~\cite{zhang2018residual}} & LTE~\cite{lee2022local} & {\fontsize{8}{10}\selectfont  28.02/0.805/0.233}  &  22.5   &  2011.3 &  0.519  \\ 
				& CLIT~\cite{chen2023cascaded}  & {\fontsize{8}{10}\selectfont  28.02/0.805/0.238}   & 37.7  &   7341.9     &1.655  \\ 
				& OPE~\cite{song2023ope}  & {\fontsize{8}{10}\selectfont 27.90/0.798/0.242}  &  22.1 &  1003.7 &  0.266  \\ 
				& GaussianSR~\cite{hu2025gaussiansr} & {\fontsize{8}{10}\selectfont 27.97/0.801/0.240}  & 23.2 &  1576.4 & 0.712 \\
				& ContinuousSR~\cite{peng2025pixel}  & {\fontsize{8}{10}\selectfont 28.04/0.805/0.236}  &  26.0 &  1980.1  &  0.319   \\ \hline 
				\multirow{5}{*}{SwinIR~\cite{liang2021swinir}} & LTE~\cite{lee2022local}    & {\fontsize{8}{10}\selectfont 28.09/0.806/0.231}   &   12.1  & 1692.8      & 0.729         \\ 
				& CLIT~\cite{chen2023cascaded} & {\fontsize{8}{10}\selectfont  28.10/0.806/0.237}   & 27.3         & 7022.3     & 1.928  \\ 
				& OPE~\cite{song2023ope} & {\fontsize{8}{10}\selectfont 28.02/0.802/0.240} &   11.7   & 684.0          & 0.438     \\ 
				& GaussianSR~\cite{hu2025gaussiansr}   & {\fontsize{8}{10}\selectfont 28.06/0.804/0.236}  &  12.8  &  1257.9 &  0.923  \\
				& ContinuousSR~\cite{peng2025pixel}  & {\fontsize{8}{10}\selectfont 28.09/0.807/0.233}  &   15.6 &  1661.6  &  0.502 \\ 
				\hline 
				\multicolumn{2}{c|}{VideoINR~\cite{chen2022videoinr}}   &  {\fontsize{8}{10}\selectfont 27.89/0.802/0.221}  &  11.3    & 1676.5    & 0.676   \\ 
				\multicolumn{2}{c|}{MoTIF~\cite{chen2023motif}}   & {\fontsize{8}{10}\selectfont 28.02/0.810/0.219}    & 12.6       & 2826.2    & 1.132      \\ 
				\multicolumn{2}{c|}{BF-STVSR~\cite{kim2025bf}}  &  {\fontsize{8}{10}\selectfont 28.14/0.812/0.213}  &  13.5     &  1876.4  & 1.003 \\ 
				\multicolumn{2}{c|}{SAVSR~\cite{li2024savsr}}   &  {\fontsize{8}{10}\selectfont 29.67/0.849/0.193}  & 11.5     &  1148.0 &  0.817 \\ 
				\multicolumn{2}{c|}{ST-AVSR~\cite{shang2024arbitrary}}    & {\fontsize{8}{10}\selectfont 29.70/0.852/0.195}    &   27.9    &  \textbf{296.8}  &  \textbf{0.101}    \\  
				
				\hline
				\multicolumn{2}{c|}{BasicAVSR (Ours)}    & {\fontsize{8}{10}\selectfont  \textbf{29.98}/\textbf{0.857}/\textbf{0.188}}     & \textbf{6.2}       & 331.2      & 0.116   
				\\ 
			\end{tabular}
			%}
	\end{table*}
	
	\subsubsection{Comparison on REDS}
	Benefiting from the pixel-level motion compensation and the multi-scale frequency prior,
	our BasicAVSR achieves the best results under all evaluation metrics and across all scaling factors, presented in Table~\ref{table:reds}. 
	As can be observed from the table, ContinuousSR demonstrates the best overall performance among AISR methods. However, it generates striped artifacts when handling scale factors of 2 and below, so its performance metrics for $\times 2$ are not included in the table. AVSR methods like SAVSR and ST-AVSR significantly outperform existing AISR methods, underscoring the importance of temporal modeling for video restoration tasks. SAVSR employs a bidirectional RNN within a window, which not only limits its performance but also impacts algorithmic efficiency (as shown in the efficiency comparison in Table~\ref{table:gen_gopro}). ST-AVSR leverages long-sequence modeling and structural and textural priors to achieve better reconstruction results. Our BasicAVSR further improves the accuracy of reconstructed details based on ST-AVSR. In comparison with ST-AVSR, our BasicAVSR achieves approximately 0.5 to 1.7 dB PSNR gains in super-resolution across various scaling factors.
	As illustrated in Fig.~\ref{fig:reds}, compared to AISR methods, AVSR methods yield more satisfactory visual results. The dramatic visual quality improvements can also be clearly seen in Fig.~\ref{fig:reds}, in which BasicAVSR recovers more faithful detail with less severe distortion across different scales. For example, the numbers on license plates are reconstructed more clearly, and the patterns on clothing in extreme video super-resolution (at a scale factor of $\times 8$) are restored with more naturally delineated edges.
	\begin{figure*}[!t]\footnotesize
		\centering
		\setlength{\abovecaptionskip}{3pt} 
		\setlength{\belowcaptionskip}{0pt}
		\begin{tabular}{cccccc}
			\includegraphics[width=\linewidth]{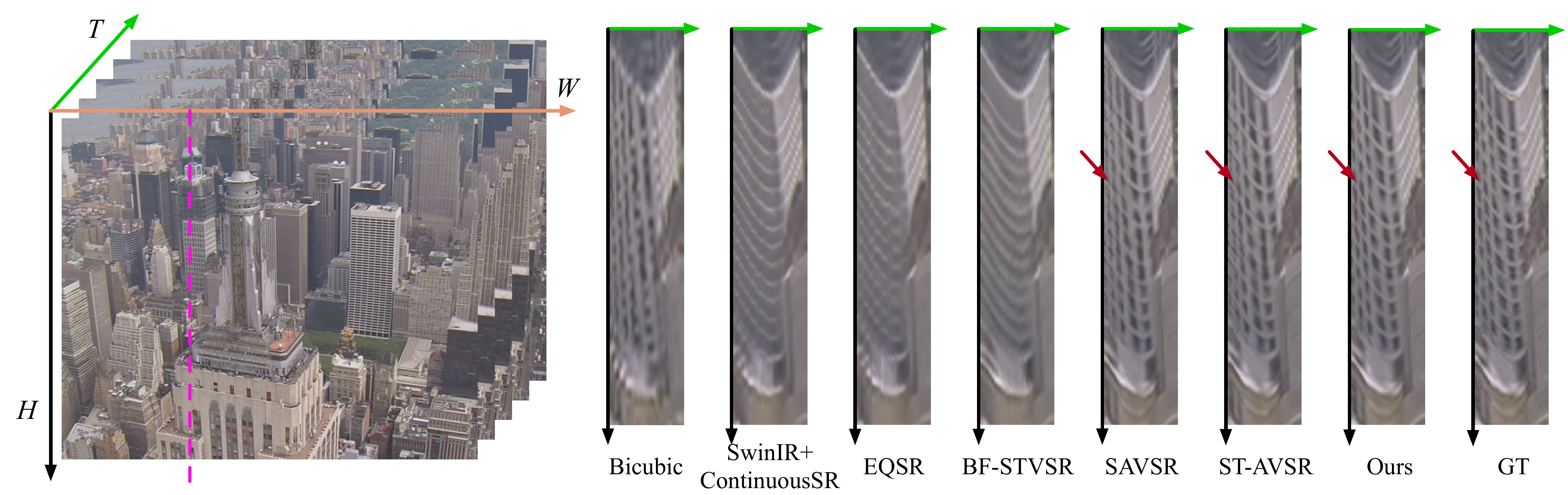}\\
		\end{tabular}
		\caption{
			Temporal consistency comparison.
			We visualize the pixel variations in the column indicated by the pink dashed line along the temporal dimension. 
		}
		\label{fig:temporal}
	\end{figure*}
	
	\subsubsection{Generalization on Vid4}
	All models trained on REDS are directly applicable to Vid4, which serves as a generalization test.
	The quantitative results, listed in Table~\ref{table:vid4}, indicate that 
	BasicAVSR surpasses all competing methods by wide margins in terms of PSNR, SSIM, and LPIPS  across varying scaling factors. A closer look is provided in Fig.~\ref{fig:plot}, illustrating the PSNR and LPIPS variations for different scaling factors. 
	It is evident that MoTIF fails to achieve satisfactory SR performance for non-integer and asymmetric scales. This issue is mainly due to the pixel misalignment between the SR frames and ground-truth frames,  leading to oscillating PSNR values. Such oscillation is less pronounced in terms of LPIPS as it offers some degree of robustness to misalignment through the VGG feature hierarchy. As for ST-AVSR, it degrades gracefully with increasing scaling factors, including non-integer and asymmetric ones. Our BasicAVSR significantly boosts the generalization of ST-AVSR, achieving up to 1.8 dB PSNR gains in super-resolution across various scaling factors compared to ST-AVSR.
	\begin{figure*}[t]\footnotesize
		\centering
		\setlength{\abovecaptionskip}{3pt} 
		\setlength{\belowcaptionskip}{0pt}
		\begin{tabular}{cccc}
			\includegraphics[width=0.7\linewidth]{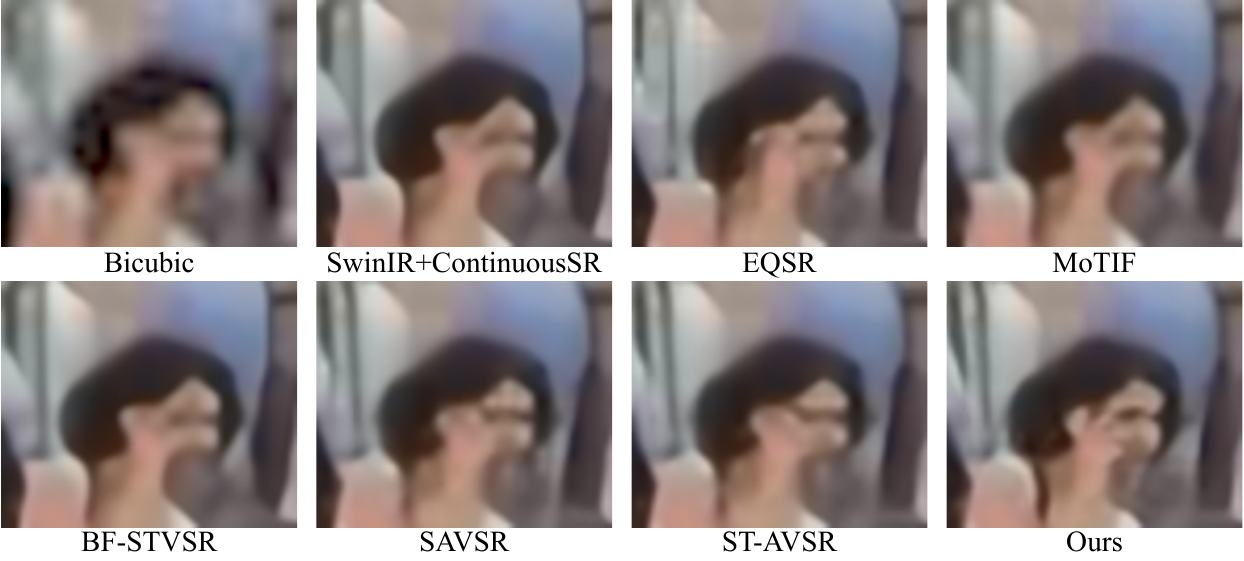}\\
		\end{tabular}
		\caption{Visual comparison of different AVSR methods under an unseen degradation model. 
		}
		\label{fig:degrad}
	\end{figure*}
	
	Qualitative results are shown in Fig.~\ref{fig:vid4}, where we find that BasicAVSR 
	consistently produces natural and visually pleasing SR outputs. 
	Particularly in high scale factor scenarios, the improvement is pronounced. As shown in the last row of Fig.~\ref{fig:vid4}, our method can accurately reconstruct letters, while ST-AVSR fails to recover fine details in the calendar. The refined method, with better alignment accuracy and the strong temporal modeling of bidirectional RNN, is more adept at reconstructing both non-structured and structured textures. 
	Additionally,  Fig.~\ref{fig:temporal} compares temporal consistency by unfolding one column of pixels as indicated by the pink dashed line along the temporal dimension. The temporal profiles of the competing methods appear blurry and zigzagging, indicating temporal flickering artifacts. In contrast, the temporal profile of BasicAVSR is closer to the ground-truth, with a sharper and smoother visual appearance.
	\begin{table*}[!t]
		\centering
		\caption{Analysis of BasicAVSR variants across diverse scenarios on REDS (PSNR$\uparrow$ / SSIM$\uparrow$ / LPIPS$\downarrow$). }
		\renewcommand{\arraystretch}{1.2}
		\label{table:extend}
		\resizebox{\linewidth}{!}{
			\begin{tabular}{c|c|c|c|c|c|c|c}
				% \toprule
				\multirow{2}{*}{Model} & \multirow{2}{*}{Online} & \multicolumn{5}{c|}{Scale} &  {Inference}\\  
				% \hline
				\cline{3-7}
				& & {$\times 2$} & {$\times 3$} & {$\times 4$} & {$\times 6$} & {$\times 8$} & Time (s) \\ 
				\hline %\hline 
				{Unidirectional RNN}  & \checkmark & {\fontsize{8}{10}\selectfont 36.20/0.964/0.046} & {\fontsize{8}{10}\selectfont 32.55/0.926/0.078} & {\fontsize{8}{10}\selectfont 30.27/0.882/0.131}&{\fontsize{8}{10}\selectfont 27.30/0.794/0.242}& {\fontsize{8}{10}\selectfont 25.55/0.727/0.315}  & 0.050\\
				{Unidirectional RNN with Lookahead (L=1)}  & \ding{55}   & {\fontsize{8}{10}\selectfont 36.84/0.968/0.043} & {\fontsize{8}{10}\selectfont 33.43/0.937/0.067} & {\fontsize{8}{10}\selectfont 31.09/0.899/0.112}&{\fontsize{8}{10}\selectfont 27.96/0.815/0.220}& {\fontsize{8}{10}\selectfont 26.10/0.749/0.296} & 0.073\\
				{Unidirectional RNN with Lookahead (L=2)}  & \ding{55}   & {\fontsize{8}{10}\selectfont 36.98/0.969/0.042} & {\fontsize{8}{10}\selectfont 33.62/0.940/0.064} & {\fontsize{8}{10}\selectfont 31.25/0.902/0.109}&{\fontsize{8}{10}\selectfont 28.09/0.819/0.217}& {\fontsize{8}{10}\selectfont 26.21/0.753/0.293}&  0.096\\
				{Unidirectional RNN with Lookahead (L=3)}  & \ding{55}    & {\fontsize{8}{10}\selectfont 37.03/0.969/0.041} & {\fontsize{8}{10}\selectfont 33.71/0.941/0.063} & {\fontsize{8}{10}\selectfont 31.34/0.904/0.107}&{\fontsize{8}{10}\selectfont 28.17/0.821/0.214}& {\fontsize{8}{10}\selectfont 26.29/0.755/0.291} &  0.103\\
				%			{Unidirectional RNN with lookahead (L=5)}  & \ding{55}    & {\fontsize{8}{10}\selectfont 36.94/0.968/0.044} & {\fontsize{8}{10}\selectfont 33.40/0.937/0.070} & {\fontsize{8}{10}\selectfont 31.03/0.897/0.115}&{\fontsize{8}{10}\selectfont 28.01/0.816/0.220}& {\fontsize{8}{10}\selectfont 26.22/0.752/0.294} \\
				{Bidirectional RNN}  & \ding{55} & {\fontsize{8}{10}\selectfont 37.40/0.972/0.040} & {\fontsize{8}{10}\selectfont 34.82/0.954/0.050} & {\fontsize{8}{10}\selectfont 32.74/0.931/0.081}&{\fontsize{8}{10}\selectfont 29.45/0.864/0.167}& {\fontsize{8}{10}\selectfont 27.33/0.798/0.247} & 0.116 \\
			\end{tabular}
		}
	\end{table*}
	\begin{table*}[!t]
		\centering
		\caption{Ablation analysis of BasicAVSR on REDS (PSNR$\uparrow$ / SSIM$\uparrow$ / LPIPS$\downarrow$). See the text for the details of different variants.}
		\renewcommand{\arraystretch}{1.2}
		\label{table:ablation}
		\resizebox{\linewidth}{!}{
			\begin{tabular}{c|c|c|c|c|c|c|c|c|c|c|c|c}
				% \toprule
				\multicolumn{3}{c|}{Upsampling}  & \multicolumn{2}{c|}{Priors} & \multicolumn{2}{c|}{Alignment} &  \multicolumn{5}{c|}{Scale} & Inference \\  
				% \hline
				\cline{1-12}
				\ding{172} &  \ding{173}  &  \ding{174} &  \ding{172}  &  \ding{173}  &  \ding{172} &  \ding{173}  &  {$\times 2$} & {$\times 3$} & {$\times 4$} & {$\times 6$} & {$\times 8$}  & Time (s) \\ 
				\hline 
				
				\checkmark & \ding{55}  & \ding{55} & \checkmark & \ding{55}  & \checkmark  &  \ding{55} &   {\fontsize{8}{10}\selectfont 36.39/0.964/0.046} & {\fontsize{8}{10}\selectfont 33.35/0.938/0.069} & {\fontsize{8}{10}\selectfont 31.28/0.900/0.117}&{\fontsize{8}{10}\selectfont 28.15/0.817/0.220}& {\fontsize{8}{10}\selectfont 26.19/0.742/0.299} &  0.071  \\
				
				\ding{55}  & \checkmark & \ding{55} &  \checkmark & \ding{55}  & \checkmark &  \ding{55}  &  {\fontsize{8}{10}\selectfont 37.05/0.969/0.043} & {\fontsize{8}{10}\selectfont 34.15/0.947/0.060} & {\fontsize{8}{10}\selectfont 31.99/0.920/0.102}&{\fontsize{8}{10}\selectfont 28.79/0.844/0.193}& {\fontsize{8}{10}\selectfont 26.78/0.774/0.270} &  0.678 \\
				
				\ding{55}  & \ding{55} & \checkmark &  \checkmark & \ding{55} & \checkmark & \ding{55}   & {\fontsize{8}{10}\selectfont 37.07/0.970/0.043} & {\fontsize{8}{10}\selectfont 34.16/0.948/0.059} & {\fontsize{8}{10}\selectfont 31.98/0.919/0.101} &{\fontsize{8}{10}\selectfont 28.77/0.844/0.192}& {\fontsize{8}{10}\selectfont 26.75/0.773/0.271}   &  0.106 \\
				
				\ding{55}  & \ding{55} & \checkmark& \ding{55} &  \checkmark & \checkmark & \ding{55}  &  {\fontsize{8}{10}\selectfont 37.12/0.971/0.042} & {\fontsize{8}{10}\selectfont 34.18/0.948/0.059} & {\fontsize{8}{10}\selectfont 32.01/0.920/0.100} & {\fontsize{8}{10}\selectfont 28.80/0.845/0.192}& {\fontsize{8}{10}\selectfont 26.78/0.774/0.270}  &  0.076 \\
				
				\ding{55}  & \ding{55} & \checkmark &   \ding{55} &  \checkmark & \ding{55}  & \checkmark &  {\fontsize{8}{10}\selectfont 37.40/0.972/0.040} & {\fontsize{8}{10}\selectfont 34.82/0.954/0.050} & {\fontsize{8}{10}\selectfont 32.74/0.931/0.081}&{\fontsize{8}{10}\selectfont 29.45/0.864/0.167}& {\fontsize{8}{10}\selectfont 27.33/0.798/0.247} & 0.116   \\
				
			\end{tabular}
		}
	\end{table*}
	\begin{figure*}[t]\footnotesize
		\centering
		\setlength{\abovecaptionskip}{3pt} 
		\setlength{\belowcaptionskip}{0pt}
		\begin{tabular}{cccc}
			\includegraphics[width=0.8\linewidth]{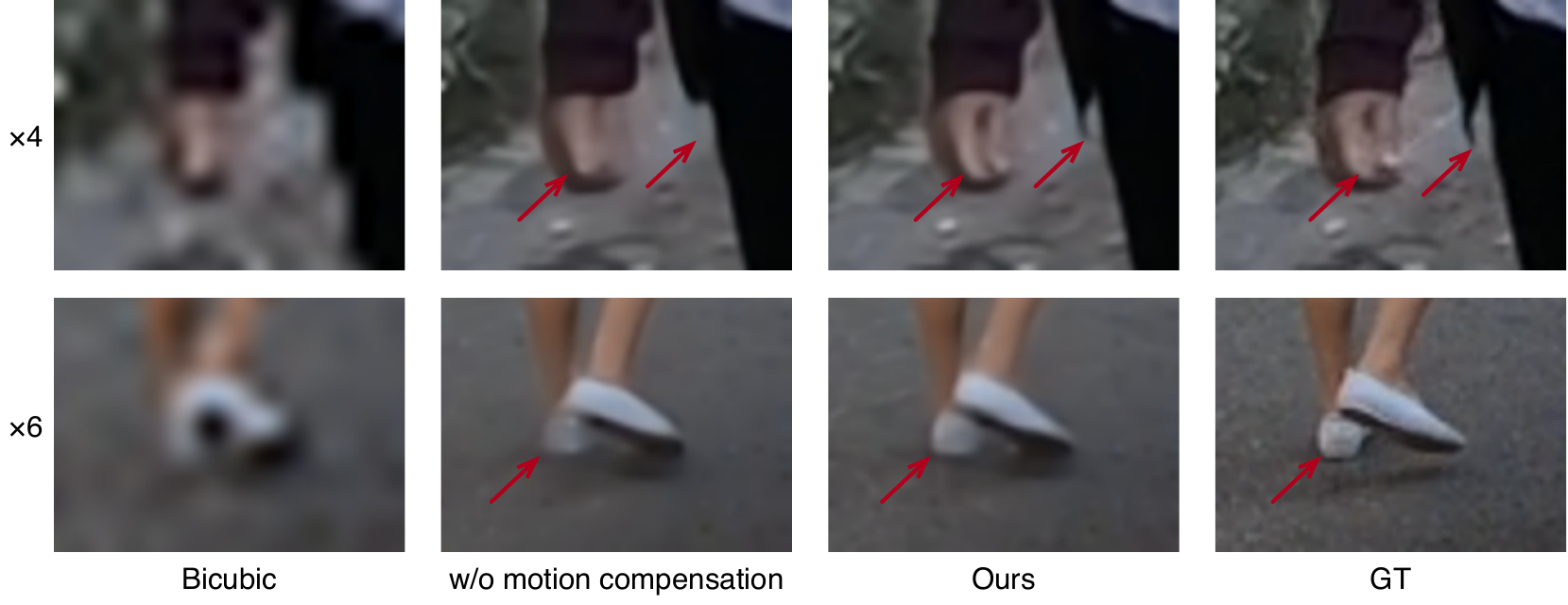}\\
		\end{tabular}
		\caption{Effectiveness of the motion compensation strategy. 
		}
		\label{fig:motion_comp}
	\end{figure*}
	\subsubsection{Generalization to Unseen Degradation Models} 
	A practical AVSR method must be effective under various, potentially unseen degradations. To evaluate this, we generate test video sequences by incorporating more complex video degradations~\cite{chan2022investigating}, such as noise and video compression before bicubic downsampling, which are absent from the training data. 
	We applied the aforementioned pipeline to the test set of GoPro to create a test set with unseen degradations. Taking $\times 4$ super-resolution as an example, the results are shown in Table~\ref{table:gen_gopro}. Our method demonstrates superior generalization compared to existing approaches. We also conducted a comprehensive comparison of all methods in terms of parameters, computational cost, and runtime using an NVIDIA RTX A6000 GPU. Our ST-AVSR and BasicAVSR not only excel in performance but are also the most efficient in processing. Our BasicVSR further improves performance without markedly increasing inference time.
	Fig.~\ref{fig:degrad} presents visual comparison of $\times 4$ SR results. 
	Due to the degradation gap between training and testing, all methods, including BasicAVSR, suffer significantly, resulting in missing details in reconstruction results. Nevertheless, BasicAVSR still produces relatively more natural and less distorted results under unseen degradations, further illustrating the superiority of our method. 
	More real-world video results are available at the link \href{https://1drv.ms/f/s!AnS8oR69EOSMiVW-tj5ZHs5qYrFI?e=1syN9v}{\textbf{\emph{Video Results}}}.

	\subsection{Adaptive Exploration for Diverse Scenarios} \label{sec:extension}
	As outlined in Section~\ref{sec:arch_exten}, we derive multiple BasicAVSR variants by adapting the propagation scheme to meet the requirements of diverse VSR scenarios.
	Table~\ref{table:extend} presents a comparative performance analysis of the different variants. 
	Despite the superior performance of bidirectional RNNs, the unidirectional variant is more flexible as it avoids storing hidden states. 
	The Unidirectional RNN with lookahead variant serves as the compromise between unidirectional and bidirectional RNNs, with its performance lying in between the two. As the value of $L$ (the number of future frames considered) increases, there is a gradual enhancement in performance, bridging the gap between the limitations of a purely unidirectional approach and the comprehensive but computationally heavier bidirectional variant.
	Compared with Tables~\ref{table:reds} and~\ref{table:gen_gopro}, all variants deliver state-of-the-art results while requiring markedly less inference time.
	These results highlight the effectiveness of our proposed core modules—including frequency priors, alignment compensation, and upsampling units—in multiple RNN architectures, underscoring their versatility and applicability across different variants.
	
	%Despite the superior performance of bidirectional RNNs, the unidirectional variants demonstrate comparable, and in some cases superior, performance in their respective application contexts, while substantially reducing computational cost. 

	\subsection{Ablation Studies} \label{sec:abla}
	In our previous work, ST-AVSR, we analyzed the necessity of several core modules. In this section, we further validate the advantages of the proposed hyper-upsampling unit, and present three variants: \ding{172} upsampling using bilinear interpolation; \ding{173} rendering RGB values pixel-by-pixel  with implicit neural representation (INR); \ding{174} our hyper-network for pre-computing the upsampling kernel.
	We also analyze the two main improved strategies in this paper, namely AVSR priors, as well as the alignment strategy. For AVSR priors, we present two variants: \ding{172} the structural and textural priors used in ST-AVSR; \ding{173} the proposed frequency priors generated by image Laplacian pyramids. As for the alignment strategy, we consider: \ding{172}  the backwarping used in ST-AVSR; \ding{173} our proposed flow compensation strategy.
	From the analysis in Table~\ref{table:ablation}, the necessity of our hyper-upsampling strategy is evident. While direct interpolation offers faster inference, it falls short in reconstructing high-quality results for super-resolution across various scaling factors. The INR achieves results comparable to ours but at a much higher computational cost. Our upsampling strategy strikes a good balance between computational cost and performance.
	Regarding AVSR priors, applying image Laplacian pyramids can also guide arbitrary-scale super-resolution effectively. Like VGG, Image Laplacian pyramids can also capture the frequency differences of objects across different scales. Replacing the original VGG with Laplacian pyramids reduces network parameters and inference time while slightly improving performance.
	For the alignment strategy, existing VSR methods mainly rely on optical flow networks. Inaccurate alignment can degrade performance. Our flow compensation strategy addresses this by searching for the most similar content near the displacement estimated by optical flow and aligning it with the current frame. As shown in Fig.~\ref{fig:motion_comp}, our compensation-based alignment method achieves more precise alignment. It effectively reduces motion artifacts in the reconstruction. Moreover, it can recover some missing details from neighboring frames, bringing the results closer to ground-truth. For instance, the hem missing in the input is successfully reconstructed by our method through alignment with adjacent frames.
	In summary, we have made improvements in two critical aspects: priors and alignment strategy. While the flow compensation strategy introduces additional computational cost, the more direct image-based prior helps to mitigate this by reducing runtime. Overall, the enhanced method achieves PSNR gains of 0.33 to 0.76 dB across various upscaling factors for super-resolution tasks, with no significant impact on the inference time.
	
	\section{Conclusion}
	
	In this paper, we present an enhanced versatile baseline for arbitrary-scale video super-resolution. 
	By rethinking the inherent limitations of current priors and alignment strategies in AVSR, we first introduce multi-scale frequency priors derived from the image Laplacian to guide arbitrary-scale video super-resolution—requiring no extra parameters and delivering both efficiency and effectiveness. We then replace the backward warping of existing methods with a second-order motion-compensation strategy for feature alignment, yielding a stronger baseline dubbed BasicAVSR.
	Our model significantly boosts video SR performance without major sacrifices in runtime efficiency, proving the effectiveness of these enhancements. 
	Moreover, we extend our method to two other versions for diverse application scenarios, with experiments confirming our strategies can be effectively applied across different scenarios. 
	Extensive experiments show BasicAVSR outperforms state-of-the-art methods in SR quality and generalization, achieving a good balance between inference speed and performance.
\ifCLASSOPTIONcaptionsoff
  \newpage
\fi



%\bibliographystyle{IEEEtran}
%\begin{thebibliography}{1}
%\end{thebibliography}

{\small
	\bibliographystyle{ieee_fullname}
	\bibliography{egbib}
}


%\end{IEEEbiography}
%% insert where needed to balance the two columns on the last page with
%%\newpage


%\vfill


\end{document}